\documentclass{edm_article}
\usepackage{microtype}

\usepackage{inconsolata}
\usepackage{array, booktabs, makecell}
\usepackage{mathtools, nccmath}
\usepackage{lipsum}
\usepackage{amsmath}
\usepackage{algorithm}
\usepackage{algorithmic}
\usepackage{listings}
\usepackage{multirow}
\usepackage{url}
\usepackage{booktabs}
\usepackage{xspace} 
\usepackage[shortlabels]{enumitem}

\usepackage{xcolor}
\usepackage{subcaption}
\usepackage{mystyle}
\usepackage{mlsymbols}
\usepackage{times}
\usepackage{amssymb} 
\usepackage{graphicx}
\usepackage{tabularx}

\definecolor{mdgreen}{rgb}{0.05,0.6,0.05}
\definecolor{darkcyan}{rgb}{0.0, 0.55, 0.55}
\newcommand{\lbl}[1]{\textsc{#1}}
\newcommand{\dataset}[1]{\textcolor{black}{\underline{#1}}}

\newcommand{\TNONE}{\textcolor{darkcyan}{\lbl{T-None}}\xspace}
\newcommand{\TKET}{\textcolor{darkcyan}{\lbl{T-KpTg}}\xspace} 
\newcommand{\TGSR}{\textcolor{darkcyan}{\lbl{T-GStuR}}\xspace} 
\newcommand{\TRES}{\textcolor{darkcyan}{\lbl{T-Restat}}\xspace} 
\newcommand{\TREV}{\textcolor{darkcyan}{\lbl{T-Revoic}}\xspace} 
\newcommand{\TPRA}{\textcolor{darkcyan}{\lbl{T-PrsAcc}}\xspace} 
\newcommand{\TPRR}{\textcolor{darkcyan}{\lbl{T-PrsRea}}\xspace} 

\newcommand{\SNONE}{\textcolor{blue}{\lbl{S-None}}\xspace}
\newcommand{\SRAS}{\textcolor{blue}{\lbl{S-RelTo}}\xspace}
\newcommand{\SAMI}{\textcolor{blue}{\lbl{S-AskMI}}\xspace}
\newcommand{\SMAC}{\textcolor{blue}{\lbl{S-MClaim}}\xspace}
\newcommand{\SPRE}{\textcolor{blue}{\lbl{S-ProEvi}}\xspace}

\newcommand{\DIALACT}[1]{\textcolor{brown}{\textit{#1}}}
\newcommand{\DAWHQ}{\DIALACT{Wh-Question}\xspace}
\newcommand{\DASNO}{\DIALACT{Statement-non-opinion}\xspace}

\newcommand{\DAYNQ}{\DIALACT{Yes-No-Questions}\xspace}
\newcommand{\DAAD}{\DIALACT{Action-directive}\xspace}
 
\newcommand{\DAAB}{\DIALACT{Acknowledgment-(Backchannel)}\xspace}
\newcommand{\DACC}{\DIALACT{Conventional-Closing}\xspace}
\newcommand{\DACSS}{\DIALACT{Continued-by-Same-Speaker}\xspace}

\newcommand{\TB}{\dataset{TalkMoves}\xspace}

\newcommand{\SAGA}{\dataset{SAGA22}\xspace}

\begin{document}

\title{Towards Actionable Pedagogical Feedback: \\A Multi-Perspective Analysis of Mathematics Teaching and Tutoring Dialogue}

\numberofauthors{7}
\author{
\alignauthor
Jannatun Naim\\
       \affaddr{University of Colorado Boulder}\\
       \affaddr{Boulder, CO, United States}\\
       \email{jannatun.naim@colorado.edu}
\alignauthor
Jie Cao\\
       \affaddr{University of Oklahoma}\\
       \affaddr{Norman, OK, United States}\\
       \email{jie.cao@ou.edu}
\alignauthor
Fareen Tasneem\\
       \affaddr{University of Chittagong}\\
       \affaddr{Chittagong, Bangladesh}\\
       \email{fareen.tasneem@gmail.com}
\and
\alignauthor
Jennifer Jacobs\\
       \affaddr{University of Colorado Boulder}\\
       \affaddr{Boulder, CO, United States}\\
       \email{jennifer.jacobs@colorado.edu}
\alignauthor
Brent Milne\\
       \affaddr{Saga Education, United States}\\
       \email{bmilne@sagaeducation.org}
\alignauthor
James Martin\\
       \affaddr{University of Colorado Boulder}\\
       \affaddr{Boulder, CO, United States}\\
       \email{james.martin@colorado.edu}
\and
\alignauthor
Tamara Sumner\\
       \affaddr{University of Colorado Boulder}\\
       \affaddr{Boudler, CO, United States}\\
       \email{sumner@colorado.edu}
}
\maketitle

\begin{abstract}
Effective feedback is essential for refining instructional practices in mathematics education, and researchers often turn to advanced natural language processing (NLP) models to analyze classroom dialogues from multiple perspectives. However, utterance-level discourse analysis encounters two primary challenges: (1) multi-functionality, where a single utterance may serve multiple purposes that a single tag cannot capture, and (2) the exclusion of many utterances from domain-specific discourse move classifications, leading to their omission in feedback. To address these challenges, we proposed a multi-perspective discourse analysis that integrates domain-specific talk moves with dialogue act (using the flattened multi-functional SWBD-MASL schema with 43 tags) and discourse relation (applying Segmented Discourse Representation Theory with 16 relations). Our top-down analysis framework enables a comprehensive understanding of utterances that contain talk moves, as well as utterances that do not contain talk moves. This is applied to two mathematics education datasets: TalkMoves (teaching) and SAGA22 (tutoring). Through distributional unigram analysis, sequential talk move analysis, and multi-view deep dive, we discovered meaningful discourse patterns, and revealed the vital role of utterances without talk moves, demonstrating that these utterances, far from being mere fillers, serve crucial functions in guiding, acknowledging, and structuring classroom discourse. These insights underscore the importance of incorporating discourse relations and dialogue acts into AI-assisted education systems to enhance feedback and create more responsive learning environments. Our framework may prove helpful for providing human educator feedback, but also aiding in the development of AI agents that can effectively emulate the roles of both educators and students.

\end{abstract}


\section{Introduction}
Research increasingly supports dialogic teaching~(learning) - an approach that encourages student-driven academic discourse — as a means of enhancing motivation and learning outcomes~\cite{boheim2021changes,howe2019teacher,resnick2018next,webb2019details}. A critical element of dialogic instruction is accountable talk, which structures classroom discourse around three key dimensions: learning community, content knowledge, and rigorous thinking~\cite{michaels2010accountable,o2019supporting}. Analyzing discourse in mathematic teaching and tutoring is essential for understanding its effectiveness.

Researchers and educators can assess dialogic teaching by examining who speaks, the nature of their contributions, and the extent to which discussions reflect multiple perspectives~\cite{lefstein2013better}. Classroom observations, whether conducted by educators or researchers, can document changes in discourse patterns over time, providing valuable insights for professional learning~\cite{calcagni2023developing}. Historically, discourse analysis relied on labor-intensive qualitative methods, including detailed human annotation of classroom interactions~\cite{mercer2019routledge}. Advances in recording technologies and natural language processing~(NLP) have introduced scalable alternatives, enabling interested parties to capture and analyze classroom dialogue more efficiently~\cite{franklin2018using,ramos2022pedagogical}.

Recent advancements in educational technology have demonstrated the significant potential of automated, data-driven feedback derived from classroom recordings in fostering teacher development.  The emergence of AI-driven tools has the potential to transform classroom discourse analysis, providing teachers with real-time, automated feedback on their instructional strategies. Improvements in machine learning, NLP, and automatic speech recognition have enabled researchers to develop models capable of detecting key discourse features, such as accountable talk~\cite{booth2024human,cao-etal-2025-enhancing,suresh2021using}, teacher questioning techniques~\cite{datta2023classifying,demszky2025automated,perkoff2023comparing},  student participation~\cite{cao2023comparative,donnelly2016automatic,wang2023can}, teacher uptakes~\cite{demszky2023m,demszky2024can}. These findings suggest that AI-driven feedback mechanisms can serve as powerful tools for enhancing interactive teaching practices across various educational settings. 
A key challenge lies in delivering more actionable, context-specific feedback to educators to enhance their instructional effectiveness. One potential solution is to provide a detailed analysis from multiple perspectives and varying levels of granularity, ensuring insights are both comprehensive and practical.

\subsection{Background and Related Work}
\label{ssec:backgroud-relatedwork}
Previous research on providing feedback to educators in multiple granularities or from a variety of perspectives often use different refined NLP models to solve multiple discourse analysis tasks on the same dialogue. This feedback  ranges from information about turn-level discourse moves to session-level quality scores. For example, M-Powering Teachers~\cite{demszky2024can,demszky2023m} has provided instructors with feedback in the context of online programming courses and 1:1 online tutoring via three metrics: uptakes of student contributions, talk time, and questioning practices. In addition, using a dataset collected by the National Center for Teacher Effectiveness~(NCTE)~\cite{demszky2022ncte} developed classifiers to annotate five turn-level indicators of effective mathematics instruction: student on task, teacher on task, student reasoning, high uptake, and focusing questions. Another recent study \cite{tran2024multi} developed models for a multi-dimensional assessment on the quality of classroom discussion based on two measures: a sentence-level Analyzing Teaching Moves discourse measure~(ATM)~\cite{correnti2021effects,correnti2015improving}, and a session-level Instructional Quality Assessment (IQA)~\cite{matsumura2008toward}. IQA scores use a 1-4 scale on 11 dimensions for modeling teacher and student contributions, and 4 of 11 dimensions' scoring depends on the ATM moves. Other multi-perspective pedagogical feedback systems like the above examples also exist. However, turn-level discourse analysis often faces two challenges: (1) \textit{Multi-Functionalities}, where a sentence may serve multiple functions which may not be captured by a single tag ~\cite{allwoodactivity,cohen1990intentions,hancher1979classification}. Designing multiple phenomena as binary indicators could solve the multi-functionalities issue by allowing a single turn to activate multiple binary indicators. (2) \textit{Many non-move utterances often get ignored}. Domain specific speech acts such as accountable talks~\cite{michaels2010accountable} and ATMs are designed to identify \textbf{high leverage} discourse moves grounded in the corresponding theories, which leads to many utterances not being covered during discourse modeling~(denoted as non-talk moves). For example, in our prior work developing classifiers for accountable talk moves, more than 50\% of teachers' and students' utterances were classified as non-talk moves~\cite{cao-etal-2025-enhancing}, and these non-talk moves have not been well-studied~\cite{jacobs2022promoting,suresh2021using}.

\subsection{Current Study and Contribution}
\label{ssec:contribution}
 Our work also falls in the multi-perspective feedback research, but specially focus on addressing the issue of multi-functionalties and understanding the nature of non targeted moves in utterance-level code classification~(such as talk moves, ontask, and ATM, etc). Besides domain-specific talk moves (based on Accountable Talk Theory~\cite{michaels2010accountable,o2019supporting} over 12 moves), our multi-perspective feedback use a version of broad-coverage dialogue acts designed for multi-functionalities, which is based on flattened multi-functional SWBD-MASL schema~\cite{jurafsky1997switchboard} with over 43 tags. We also propose to use graph-based discourse relation parsing~(based on Segmented Discourse Representation Theory (SDRT)\cite{asher2003logics} over 16 relations) to naturally model the dependency relations among arbitrary utterance pairs. We separate out the global session-level analysis~(such as IQA or MQI) to solely focus on fine-grained discourse analysis. 
 Our contributions can be summarized as follows:

\begin{itemize}
    \item We utilize two datasets \TB~(teaching) and \SAGA(tutoring) annotated with 3-view discourse analysis: (1) domain specific talk moves, (2) dialogue acts, and (3) discourse relations to investigate pedagogical behaviors in mathematics education. We are interested in both classroom teachers and tutors because both work directly with students to support their mathematics learning. By using our proposed toolkits to automatically annotate the above three discourse views with the state-of-the-art models, we could easily conduct a comparative study on the pedagogical dynamics between teaching and tutoring domain. 
    \item We propose a top-down analysis framework to provide a thorough understanding of utterances that contain talk moves, as well as utterances that do not contain talk moves. Our methodology involved distributional unigram analysis, documenting frequent pedagogical behaviors via sequential talk move analysis, and highlighting the intersection between talk moves/non-talk moves, dialogue acts and discourse relations. Dialogue acts offer insights into the role of non-talk move utterances, while discourse relations highlight the interplay between talk and non-talk moves within the discourse flow. This integrated analysis of talk moves, dialogue acts, and discourse relations helps address the research gap on the contributions of non-talk moves to the overall discourse structure. This multi-perspective framework offers more comprehensive explainations for providing feedback to human educators and for guiding the design of AI agents to mimic the role of educators and students. 
\end{itemize}

\begin{table*}[!htb]
\centering
\renewcommand{\arraystretch}{1.2}
\setlength{\tabcolsep}{6pt}
\caption{\label{tbl:overall-statistics} Summary of Datasets on Mathematic Teaching~(\TB) and Tutoring(\SAGA)}
\begin{tabular}{lcccccc}
\hline
\textbf{Dataset} & \textbf{Sessions} & \textbf{T-Utterances} & \textbf{S-Utterances} & \textbf{Domain} & \textbf{Students per Session} & \textbf{Session Length} \\ 
\hline
\TB   & 567  & 174,168  & 59,823  & Mixed-Teaching & ~20  & 30-55 min \\
\SAGA & 121  & 33,695   & 11,115  & High School-Tutoring & 2-5  & ~35 min \\ 
\hline
\end{tabular}
\end{table*}

\begin{figure*}[!htb]
    \centering
    \Description{This is an image of a diagram illustrating a running example of classroom dialogue with associated utterance-level talk moves, dialogue acts, and discourse relations between paired interactions.}
    \includegraphics[width=1\textwidth]{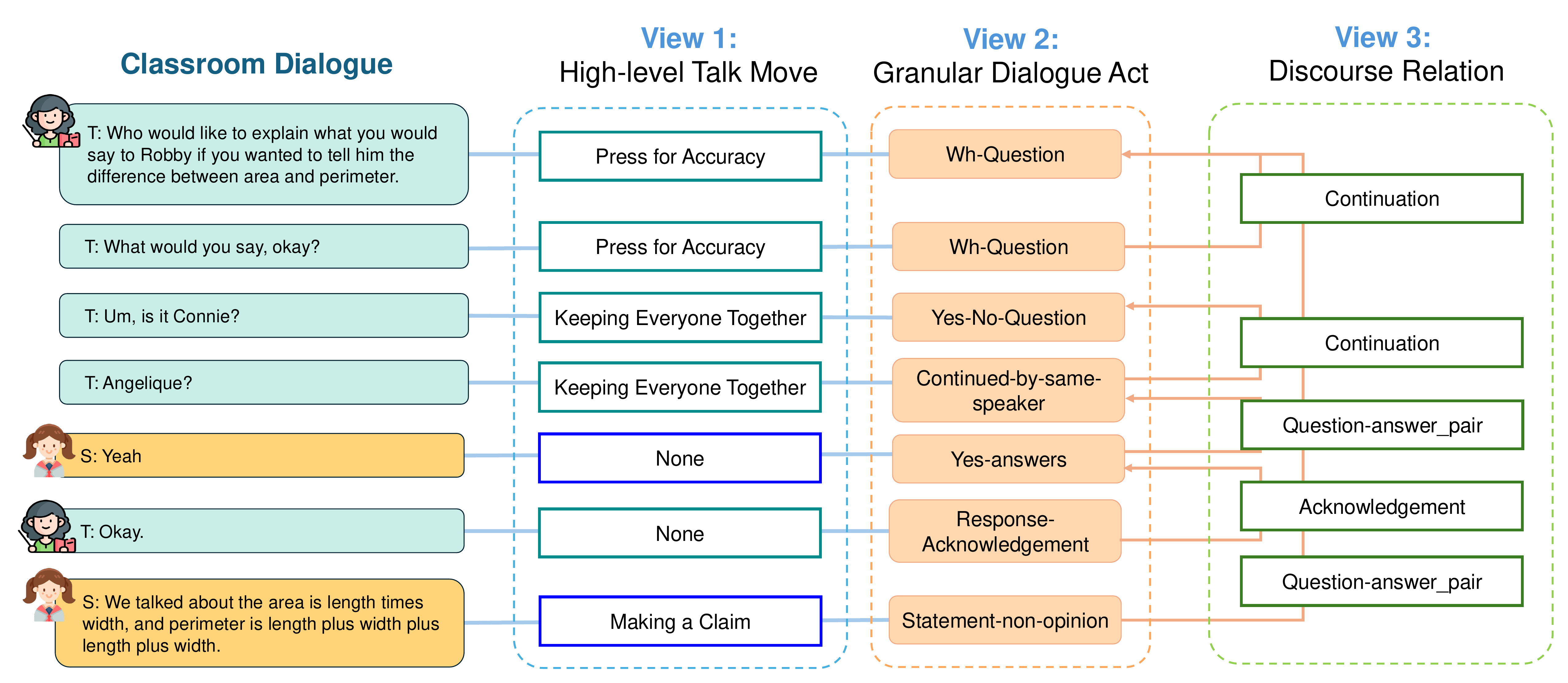} 
    \caption{A running example for our multi-perspective analysis with talk moves, dialogue acts, and discourse relations.}
     \label{fig:framework}
\end{figure*}
\section{Methods}
\subsection{Dataset}
We utilized two primary datasets from mathematics K-12 educational contexts, one teaching dataset \TB and one tutoring dataset \SAGA, summarized in Table~\ref{tbl:overall-statistics}. Each session's transcript from the two datasets has been meticulously annotated by human experts, labeling instances of 7 distinct teacher talk moves and 5 student talk moves. The \TB dataset, originally introduced by \cite{suresh2022talkmoves}, comprises 567 mathematics classroom sessions spanning a diverse range of topics across elementary to high school levels. The \SAGA dataset, originally introduced by \cite{cao-etal-2025-enhancing}, is derived from a high school tutoring dataset collected in 2022 in collaboration with Saga Education, a non-profit provider of  tutoring services. Saga partners with school districts serving low-income and historically marginalized communities to offer high dosage mathematics tutoring. Different from classroom teaching, Saga's tutoring model operates in a hybrid format, where students participate in tutoring sessions within a classroom while paraprofessional tutors engage remotely using technology. The annotated Saga dataset includes 121 sessions, totaling 69.7 hours of video, with 33,695 teacher utterances and 11,115 student utterances labeled with talk moves. 

\subsection{Multi-Perspective Dialogue Analysis}
Our approach integrates three perspectives of analysis with talk moves, dialogue acts, and discourse relations, as shown in Figure ~\ref{fig:framework}:
\paragraph{Talk Moves} Accountable Talk theory~\cite{michaels2013accountable} encompasses various types of talk moves, each differing in usage frequency and application across classroom contexts.The TalkMoves and Saga datasets are annotated for seven teacher Talk Moves~(with prefix "T" and in \textcolor{darkcyan}{dark cyan}) designed to foster student engagement, guide discussions, and facilitate productive learning interactions: keeping everyone together~(\TKET), encouraging participation~(\TGSR), restating responses for clarity~(\TRES), revoicing ideas to highlight significance~(\TREV), pressing for reasoning~(\TPRR), pushing for accuracy~(\TPRA), and an unspecified category for utterances that do not fit these moves~(\TNONE). Similarly, the datasets are annotated for five student Talk Moves~(with prefix "S" and in \textcolor{blue}{blue}) that capture how students actively contribute to discussions: making claims~(\SMAC), providing evidence and reasoning~(\SPRE), reacting to and building on peers' ideas~(\SRAS), requesting clarification~(\SAMI), and an unspecified category for utterances outside these moves~(\SNONE).This set of talk moves closely corresponds to accountable talk theory but is not exhaustive; there are other important talk moves that were not included due to their low frequency in the dataset and/or minimal extant literature~\cite{jacobs2022promoting}. This study begins with an analysis of both teacher/tutor and student talk moves. Grounded in Accountable Talk theory, we aim to deepen our understanding of classroom discourse dynamics, the effectiveness of instructional strategies, and the role of structured dialogue in enhancing student learning and engagement. View 1 in Figure~\ref{fig:framework} shows the corresponding talk moves for each utterance in the example classroom dialogue. In our analysis, we used the human annotated talk moves. For future analysis, existing toolkits for modeling talk moves that include pretrained language models could be used~\cite{suresh2021using,suresh2022fine,10.1145/3657604.3664664,cao-etal-2025-enhancing}.
\paragraph{Dialogue Act} Talk moves can be viewed as a set of domain specific dialogue acts~(DAs) tailored for the educational domain. The foundational idea that each utterance performs an action was first introduced by philosopher Wittgenstein~\cite{wittgenstein2010philosophical}. Speech act theory, developed by Austin~\cite{austin1975things} and later expanded by his student Searle~\cite{searle1969speech}, was one of the earliest frameworks for categorizing communicative actions. DAMSL (Dialogue Act Markup in Several Layers)\cite{allen1997draft, core1997coding} was proposed to address the issue of multi-functionalities~\cite{allwoodactivity,cohen1990intentions,hancher1979classification} by allowing utterances to serve multiple roles across independent layers, such as \emph{Communicative-Status}, \emph{Information-Level}, \emph{Forward-Looking Function}, and \emph{Backward-Looking Function}. Each layer contains multiple subcategories, resulting in a rich but complex tagging scheme.  A major challenge with DAMSL is the vast number of possible tag combinations, which complicates both manual and automatic annotation.  To mitigate this, SWBD-DAMSL~\cite{jurafsky1997switchboard}, an adaptation of DAMSL for the Switchboard Corpus~\cite{godfrey1992switchboard}, was introduced. It reduces the tag set to 220 combinations (including 28 new tags not in DAMSL) and clusters them into 42 mutually exclusive categories, thereby flattening the multi-functional labels. This reduction significantly improves the feasibility of automatic dialogue act annotation and the development of dialogue-act-specific language models for speech recognition~\cite{stolcke2000dialogue}, while sacrificing the flexibility and expressiveness in capturing multi-functionality. Comparing to the talk moves in View 1, View 2 in Figure~\ref{fig:framework} demonstrates the more finegrained DAs for each utterance. In our work, we follow SWBD~-DAMSL's 42 tags~\footnote{\url{https://web.stanford.edu/~jurafsky/ws97/manual.august1.html}} with an extra tag "+", which means continued previous talk by the same speaker. Among those top-tier models, we mainly considered the open-sourced models as possible toolkits. The final model we selected is a hierarchical neural architecture with a Bi-GRU on top of trainable speaker-aware utterance encodings from RoBERTa-base~\cite{liu2019roberta}, which jointly tags 196 utterances in a chunk window into their corresponding DAs~\cite{he-etal-2021-speaker-turn}. We pretrained a model on the SWBD corpus with 43 tags~(such as \DAWHQ, \DASNO, \DAYNQ etc, denoted in italic and the color \textcolor{orange}{orange}), obtaining 82.4 accuracy, which is approaching the state-of-the-art models with 83.1 on 42 tags. The model results are replicable with its open-sourced code on the github~\footnote{\url{https://github.com/zihaohe123/speak-turn-emb-dialog-act-clf}}. 

\paragraph{Discourse Relation} Beyond the utterance-level talk moves and dialogue acts, discourse relations as a structural dependency is able to capture the multi-functional nature of an utterance in relation to its neighboring utterances. It has been extensively studied across various discourse theories~\cite{benamara-taboada-2015-mapping,fu-2022-towards,jon-dda2022}, such as Rhetorical Structure Theory (RST)\cite{mann1988rhetorical}, Discourse Representation Theory (DRT)\cite{kamp2011discourse}, Hobbs' theory of discourse~\cite{hobbs1990literature}, and the Penn Discourse Treebank (PDTB) framework~\cite{prasad2008penn}. Among those, Segmented Discourse Representation Theory (SDRT)\cite{asher2003logics}, an extension of DRT, offers a hierarchical model of text organization with full discourse annotation. Several corpora, including DISCOR\cite{reese2007reference}, ANNODIS~\cite{afantenos2012empirical} and STAC~\cite{asher-etal-2016-discourse}, implement SDRT using directed acyclic graphs (DAGs) that allow multiple parent nodes but prohibit crossing edges. The success of SDRT has made it as a popular framework to study discourse relations in various multiparty dialogues, such as Molweni for Ubuntu online forum~\cite{li-etal-2020-molweni} and Minecraft Structured Dialogue Corpus~(MSDC) for jointly modeling conversation moves and builder moves~\cite{thompson-etal-2024-discourse}. In this work, we mainly focused on the SDRT analysis for mathematics dialogues. As shown in View 3 in Figure~\ref{fig:framework}, beyond the utterance-level (Views 1 and 2), the SDRT provides the labeled discourse relations between pairs of utterances if one exists. Richer expressiveness often indicates harder annotation for both human and models. Among the top-tier SDRT parsing models~\cite{bennis-etal-2023-simple,chi-rudnicky-2022-structured,li-etal-2024-discourse,thompson-etal-2024-llamipa}, we select the state-of-art model Llamipa~\cite{thompson-etal-2024-llamipa} on automatic discourse parsing. Llamipa is a LLM~(Llama3-8B) finetuned on  Minecraft Structured Dialogue Corpus~(MSDC), which has shown good generalization on MSDC~(79.51 F1) and STAC~(77.96 F1) datasets. The model we used is the public checkpoint on the huggingface model hub~\footnote{\url{https://huggingface.co/linagora/Llamipa}}. As discussed in~\S\ref{ssec:limitations}, finetuning the model with future in-domain annotation may further improve the performance.




\subsection{Top-Down Analysis Framework}
\label{ssec:top-down-framework}
We propose a 3-stage top-down framework to provide a thorough analysis of the mathematics teaching and tutoring datasets. We cluster \TNONE ~and \SNONE as non-talk moves utterances throughout our analysis. For each of the following stages, we always started with an analysis with all talk moves, then we specifically studied the non-talk moves utterances.

\subsubsection{High-level Analysis via Unigram}
\label{sssec:method-unigram}
Since our datasets include manually annotated talk moves, beginning with a faithful unigram talk move analysis ensures a reliable high-level examination of discourse patterns. To achieve a more granular understanding, we further analyzed the DAs in conjunction with the talk moves. Specifically, we examined the top three DAs corresponding to each talk move and the top seven DAs for each non-talk move utterance. These thresholds were selected to ensure that at least 50\% and 75\% of the overall distribution are captured, respectively. Utterances labeled with the dialogue act \textit{Continued-by-Same-Speaker} were excluded from this analysis, as they primarily function as extensions of preceding DAs and do not contribute meaningful standalone insights. All of our analyses compare the teaching and tutoring domains to identify similarities, differences, and domain-specific discourse patterns.


\subsubsection{Behaviors Discovered via Sequential Analysis}
\label{ssec:method-seq-analysis}
Unigram talk move analysis provides only a limited distributional perspective, making it insufficient for capturing detailed pedagogical behaviors.
To address this, we conducted a second-level analysis focusing on the sequential patterns of talk moves, including non-talk moves, to identify highly frequent interaction patterns. We only considered sequences or "transitions" with a probability of 10\% or higher between two talk moves. To better understand how different participants engaged and verbally responded, we categorized our transition analysis based on the actor (teacher or student) receiving the transition, allowing us to distinguish behavioral tendencies across different roles. We also analyzed sequences of talk moves excluding intervening non-talk moves, where we included all transitions without a probability threshold, ensuring a more comprehensive and accurate understanding of these meaningful interactions. For sequences that involved talk moves and non-talk move utterances, we analyzed the probability distribution of \TNONE~ occurring between two talk moves to capture how these \TNONE s engage with a preceding talk move and influence the transition to the succeeding one. To determine the probability, we first identified instances where the talk move pairs are separated by zero or more \TNONE~talk moves. For each possible number of \TNONE~ talk moves separating the pair, we calculated the frequency of such instances, excluding those with a frequency below 5\%. We then utilized the following equation to calculate the probability:



\begin{fleqn}
  \begin{equation}
    \begin{aligned}[b]
        & \text{Probability}_{T\text{-None}}(tm_j, tm_k) =\\
        & \begin{cases}
\frac{\sum (\text{TNoneCount}_i \times \text{Count}_i)}{\sum \text{Count}_i} \times 100, & \text{if } \sum \text{Count}_i > 0 \\
0, & \text{otherwise}
\end{cases}
    \end{aligned}
  \end{equation}
\end{fleqn}

In the above equation, $TNoneCount_i$ denotes the count of \TNONE~ utterances separating two talk moves, while $Count_i$ denotes the count of instances of two talk moves $tm_j$ and $tm_k$ separated by $TNoneCount_i$ \TNONE~ . 

\subsubsection{Deep-dive via Multi-view Analysis}
\label{sssec:method-multi-view}






Our analysis of sequential bigram pairs of talk moves provides insight into high-probability transitions that reveal key behavioral patterns of both teachers and students within classroom discourse. However, certain sequential dependencies between talk moves may remain implicit due to the presence of intervening utterances that do not contain talk moves. These non-talk moves elements can obscure direct talk move connections while still playing a crucial role in shaping the discourse dynamics. For example, \TNONE~utterances might be used by the teacher to give directions, showing students how to solve a problem, or evaluating a student's idea. To gain a deeper understanding of how utterances with and without talk moves are interwoven within these sequences and the pedagogical behavior patterns they reflect, we extend our analysis to examine the discourse relations underlying these transitions. Our investigation consists of two key components. First, we analyze direct transitions between talk move pairs to explore how they interconnect and contribute to the structured progression of discourse. Second, we examine transitions where non-talk moves utterances intervene, assessing their role in influencing dialogue flow and their potential role in shaping interactions.


\begin{figure}[ht]
    \centering
    \begin{subfigure}{0.45\textwidth}
        \Description{This is an image of a pie chart depicting the talk move ratio in the TalkMove Dataset.}
        \centering\includegraphics[width=\linewidth,trim = 30 30 30 15]{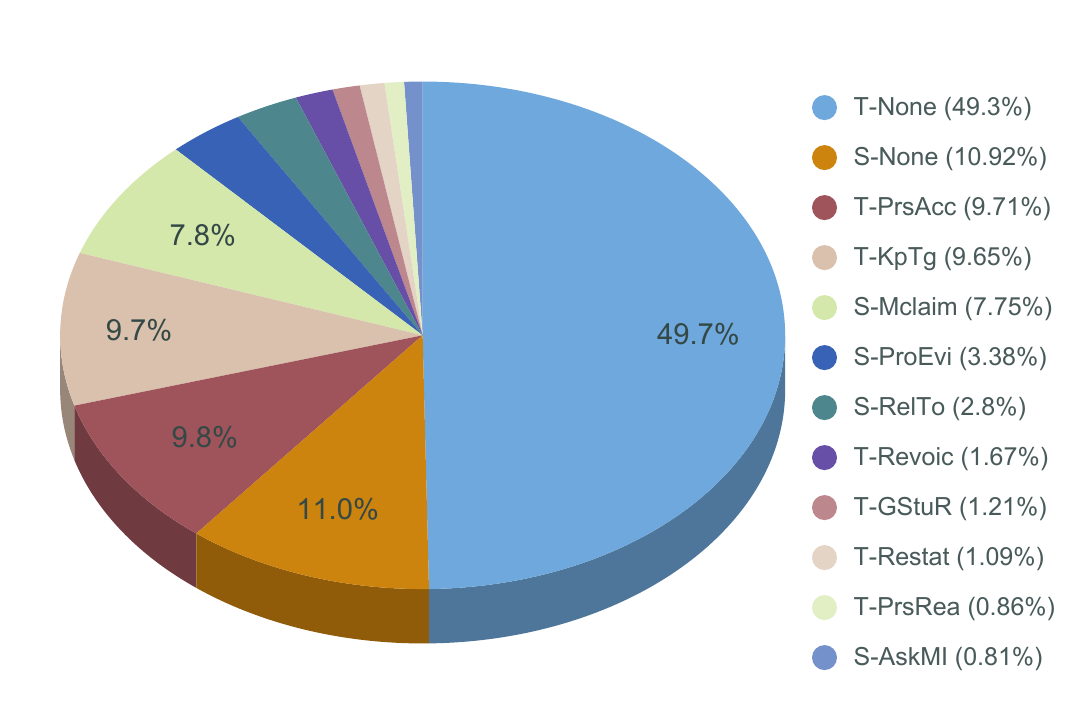}
        \subcaption{Teacher/Student Talk Moves in the TalkMove Dataset.}
        
    \end{subfigure}
    \hfill
    \begin{subfigure}{0.45\textwidth}
        \Description{This is an image of a pie chart depicting the talk move ratio in the SAGA Dataset.}
        \centering
\includegraphics[width=\linewidth, trim = 30 30 30 5]{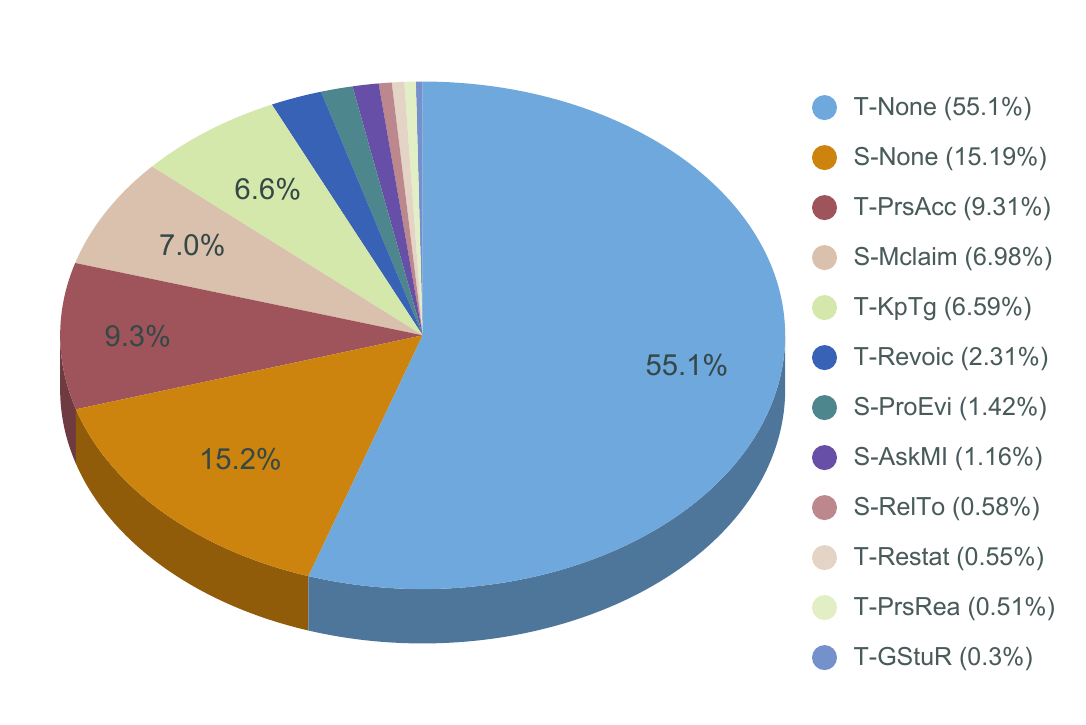}   
        \subcaption{Tutor/Student Talk Moves in the SAGA Dataset.}
        
    \end{subfigure}
    \caption{Comparison of Teacher/Tutor and Student Talk Moves across TalkMove and SAGA Dataset. }
    \label{fig:unigram}
\end{figure}

\section{Results}
\label{sec:results}
\subsection{Results from Unigram Analysis}
\label{ssec:rst-unigram-talkmvove}
\begin{figure}[!htb]
    \centering
    \Description{This is an image of a column chart illustrating the dialogue act distribution across both the \TB and \SAGA dataset.}   
    \centering\includegraphics[width=1\columnwidth]{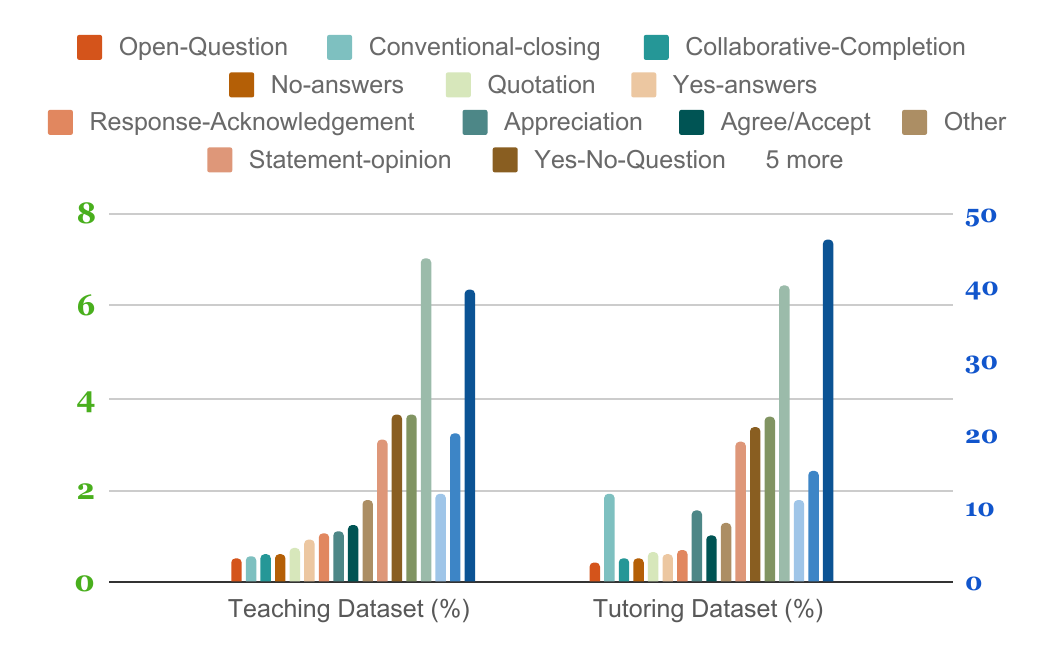} 
    \caption{Distribution of DAs in the Teaching and Tutoring Datasets. The last three DAs: \DAAD, \DASNO, and \DACSS are represented on the right scale, while all others are on the left scale. Only DAs with a ratio greater than 0.5\% are displayed.}
    \label{fig:daoverall}
\end{figure}

By examining the unigram distribution of teacher and student talk moves in both the teaching and tutoring datasets, Figure~\ref{fig:unigram} shows that the tutoring dataset contains a higher proportion of utterances without talk moves (\TNONE~and \SNONE) for both instructors (teachers/tutors) and students, with 5.8\% more for instructors and 4.3\% more for students compared to the teaching dataset. Consequently, the talk moves have a lower proportion in the tutoring dataset, except for \TREV (2.31\% > 1.67\%) and \SAMI (1.61\% > 0.81\%). To gain further insights into the pedagogical behavior patterns and students' responses, we analyzed both the talk moves and the non-talk moves utterances with DAs to capture a more granular view. Among 43 DAs, only 30 are used in our two datasets as presented in Figure~\ref{fig:daoverall}.

\begin{figure}[!htb]
    \centering
    \begin{subfigure}{0.45\textwidth}
       \Description{This is an image of a stacked column chart depicting the dialogue act distribution in each talk move in the \TB dataset.}
        \centering
        \includegraphics[width=\linewidth, trim = 30 30 30 10]{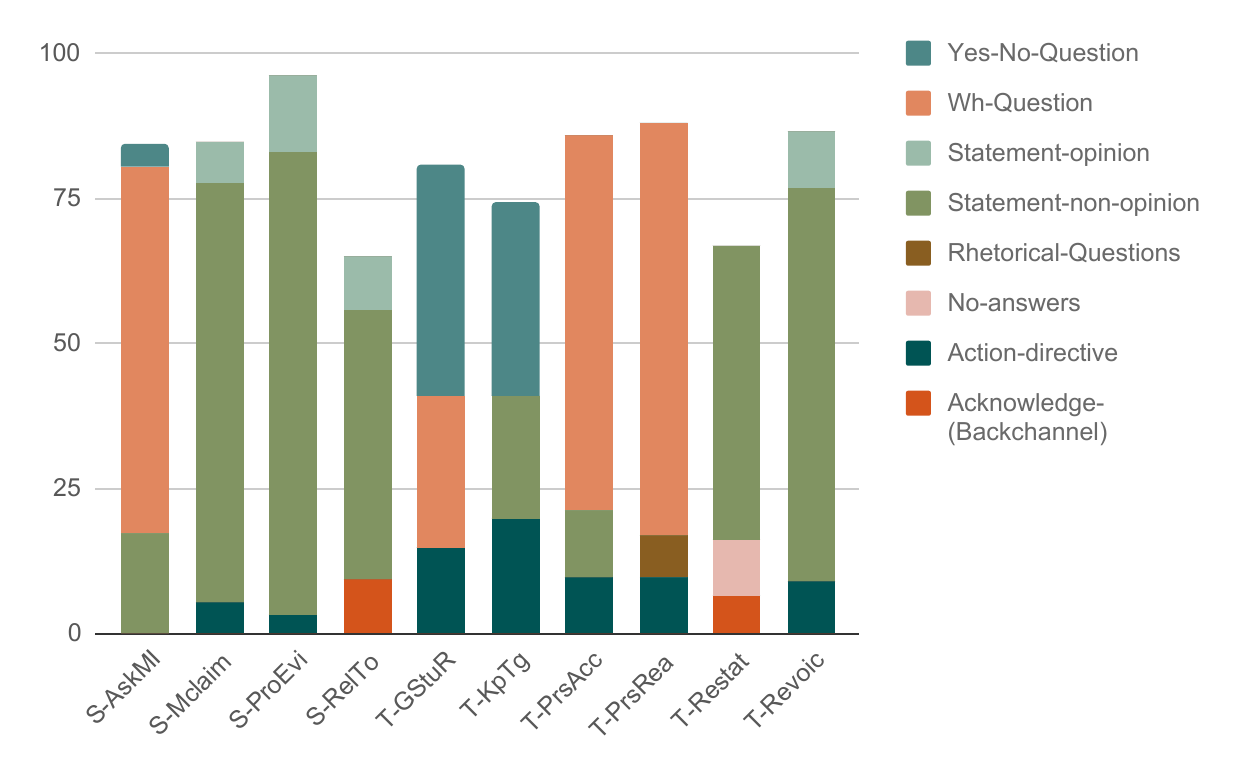}
        \subcaption{Top 3 Dialogue Acts in Talk Moves~(Teaching).}
    \end{subfigure}
    \vfill
    \begin{subfigure}{0.45\textwidth}
        \Description{This is an image of a stacked column chart depicting the dialogue act distribution in each talk move in the Saga dataset..}
        \centering
        \includegraphics[width=\linewidth, trim = 30 30 30 5]{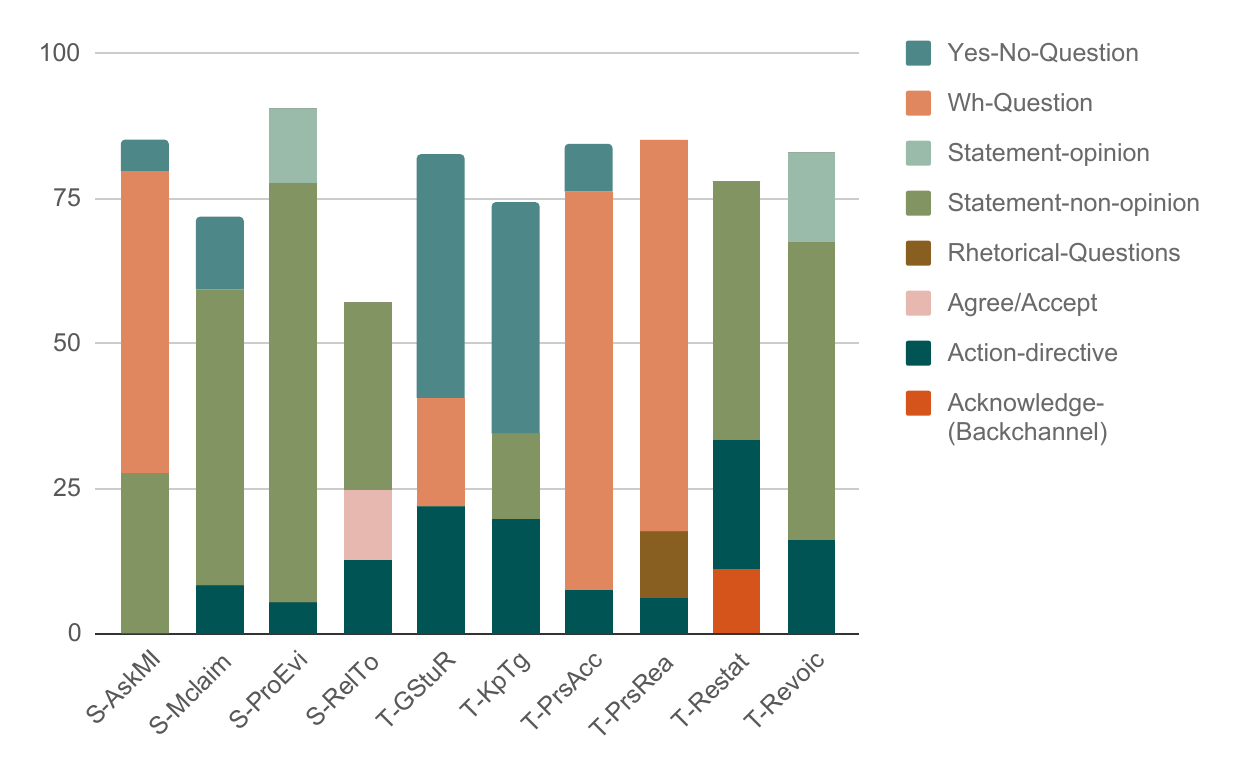}
       \subcaption{Top 3 DAs in Talk Moves~(Tutoring).}
    \end{subfigure}
\caption{Comparison of Talk Moves with DAs. }
    \label{fig:unigramda}
\end{figure}

\subsubsection{Talk Moves with DAs}
\label{ssec:core-talkmoves-da}
Our findings on talk moves with their top three associated DAs are illustrated in Figure~\ref{fig:unigramda}. As expected, \DAWHQ~naturally dominates the teacher talk moves~\TPRA, \TPRR, and the student talk move~\SAMI, which are related to asking questions. \DASNO~is the predominant DA across the student talk moves: \SMAC, \SPRE, \SRAS, and the teacher talk moves: \TRES and \TREV. \DAYNQ~is frequently used in~\TGSR, \TKET, \TPRA and~\SAMI. \DAAD is used more frequently in teacher talk moves than in student talk moves, highlighting their pivotal role in guiding and structuring classroom interactions. Besides that, students actively contribute to directing actions within classroom discussions. 

\begin{figure}[!htb]
    \centering
    \Description{This is an image showing interesting examples of teacher and student utterances with their associated talk moves and dialogue acts.}\includegraphics[width=1\columnwidth]{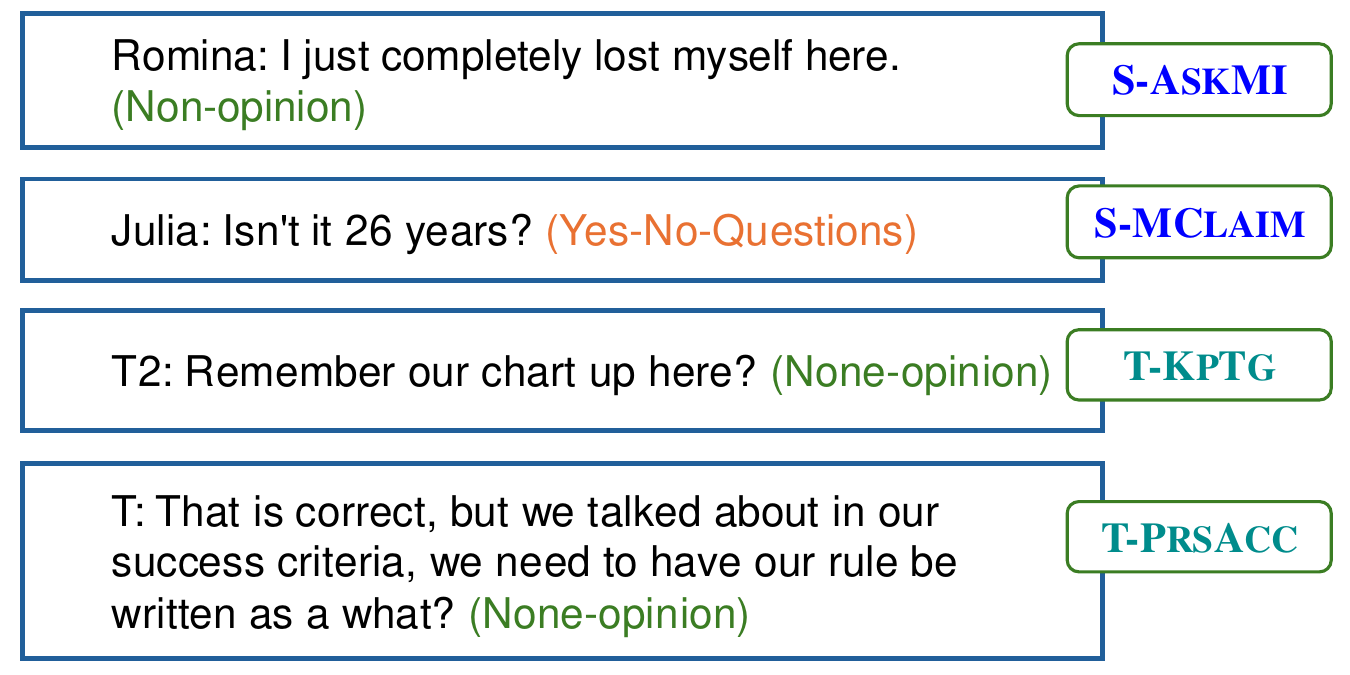} 
    \caption{Interesting DA Use Cases for Talk moves}
    \label{fig:interesting-das}
\end{figure}

\begin{figure}[!htb]
    \centering
    \Description{This is an image showing examples of teacher talk moves \TGST, \TKET, \TPRA, and the student talk moves \SAMI, \SMAC, that aligns with the dialogue act \DAYNQ.}
    \includegraphics[width=1\columnwidth]{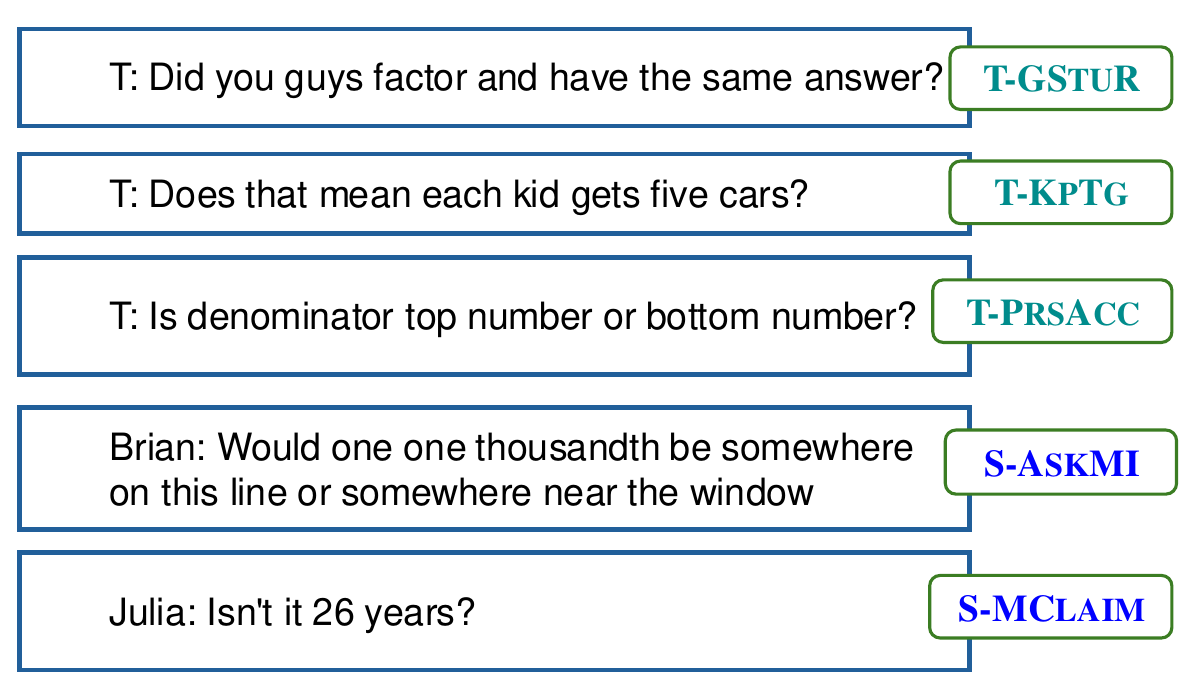} 
    \caption{Examples of teacher talk moves \TGSR, \TKET, \TPRA, and the student talk moves \SAMI, \SMAC, that aligns with the dialogue act \DAYNQ .}
    \label{fig:yesno}
\end{figure}

More interestingly, as illustrated in Figure~\ref{fig:interesting-das}, \SAMI and \TPRA typically expressed as questions, can also appear in the form of statements. For example, \SAMI being expressed as a \textit{Statement} may reflect a student behavior of implicitly conveying their need for additional information by articulating their confusion rather than making a direct request. Recognizing this phenomenon can make future agent design for mimicing and comprehending teacher or students behavior more diverse. Furthermore, \SMAC~also exhibits a notable percentage of \DAYNQ~rather than statements. Figure~\ref{fig:yesno} illustrates instances of these cases. This figure reveals that students may frame claims as \DAYNQ. This pattern suggests that students might have lower confidence when making claims. The higher occurrence of this phenomenon in the tutoring dataset, as observed from 
Figure~\ref{fig:unigramda}, could indicate an overall lower confidence level among students in the tutoring domain. Furthermore, the association of \DAYNQ with \SMAC may serve as a valuable signal for teachers or dialogue agents to intervene and support students in developing greater confidence in their responses.

Our results suggest that DAs offer a more nuanced understanding of the behavioral patterns of students and teachers embedded within high-level talk moves. When integrated into instructor feedback or the training of dialogue agents, this detailed perspective may be helpful in more accurately interpreting student needs, suggesting specific responses and enhancing engagement.

\subsubsection{Non-talk moves Interactions}
\label{sssec:non-strategic-interactions}

\begin{figure}[!htb]
    \Description{This is an image of stacked bar chart showing the comparison of \textit{None} talk moves (\TNONE~ and \SNONE) with their associated DAs.}
    
    \centering
    \includegraphics[width=1\columnwidth, trim=30 30 30 10]{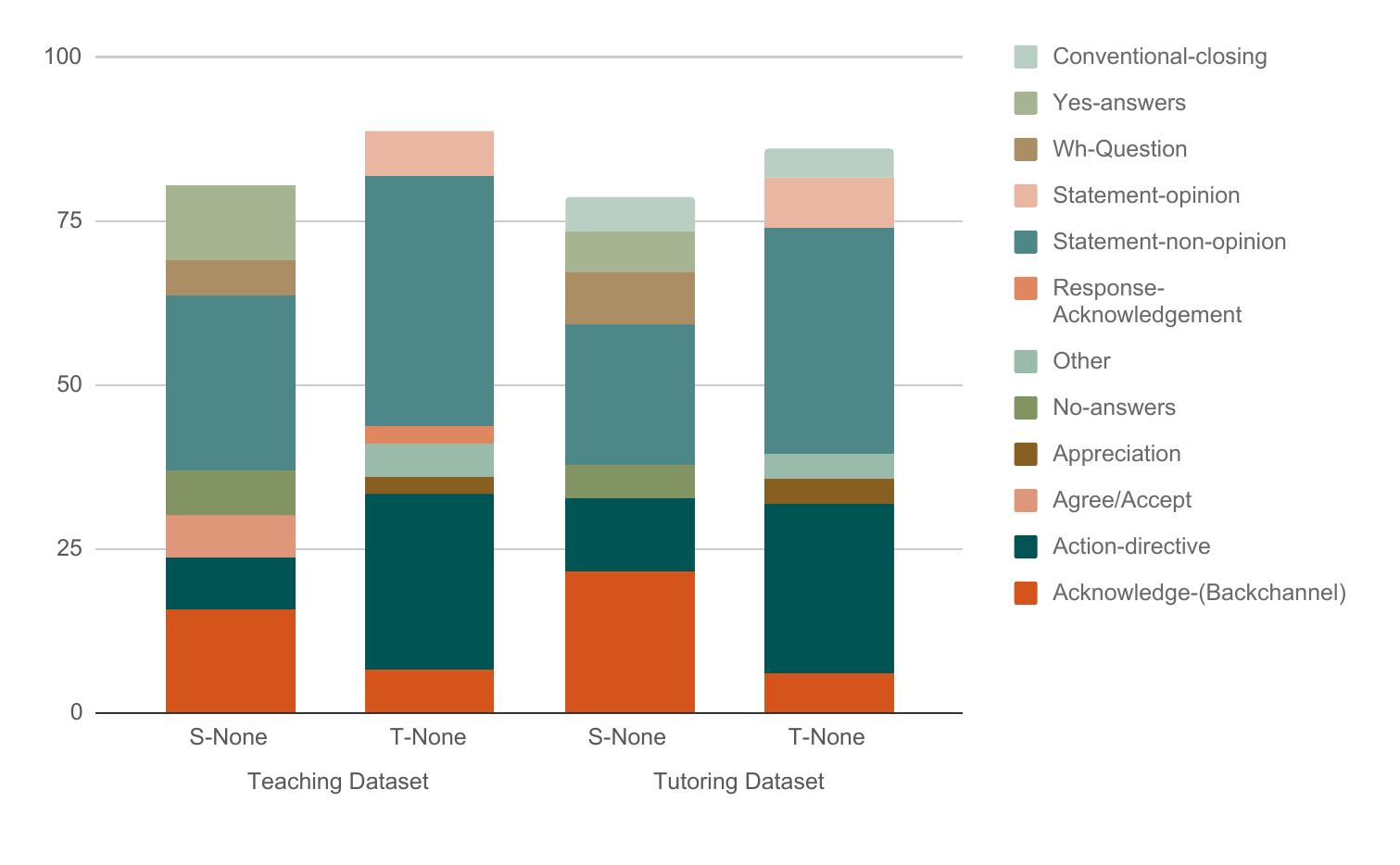} 
    \caption{Comparison of \textit{None} talk moves (\TNONE~ and \SNONE) with their associated DAs.}
    \label{fig:noneda}
\end{figure}

We also examined the utterances without talk moves (\TNONE~and \SNONE) and their associated DAs, as illustrated in Figure~\ref{fig:noneda}. Notably, both \SNONE~and~\TNONE~exhibit a similar distribution of DAs. The~\DASNO,~\DAAB and~\DAAD are the most frequent DAs in all of these utterances. Moreover, \DAAB dominates more for \SNONE than \TNONE. In the tutoring domain, \textit{None} talk moves also feature a reasonable portion of \DACC, which indicates that teachers and students are more engagingly saying "Bye" and "See you" in the online tutoring sessions.

In summary, as shown in Figure~\ref{fig:unigram}, non-talk moves utterances make up over 50\% of the dialogue exchanges in both the teaching and tutoring datasets. Analyzing these utterances complements the study of talk moves by looking at the role of a broader range of DAs that can be used to define the conversational dynamics. 

\subsection{Results on Sequential Talk Move Analysis}
\label{ssec:rst-seq-talkmove}

\subsubsection{Transition Diagrams of Talk Moves}
\label{sssec:rst-trans}

\begin{figure}[!htb]
    \centering
    \begin{subfigure}{0.47\textwidth}
        \centering
        \Description{This is an image of transition diagram depicting the transition probability between two teacher talk moves.}
            \centering
            \includegraphics[page = 3, width=\columnwidth]{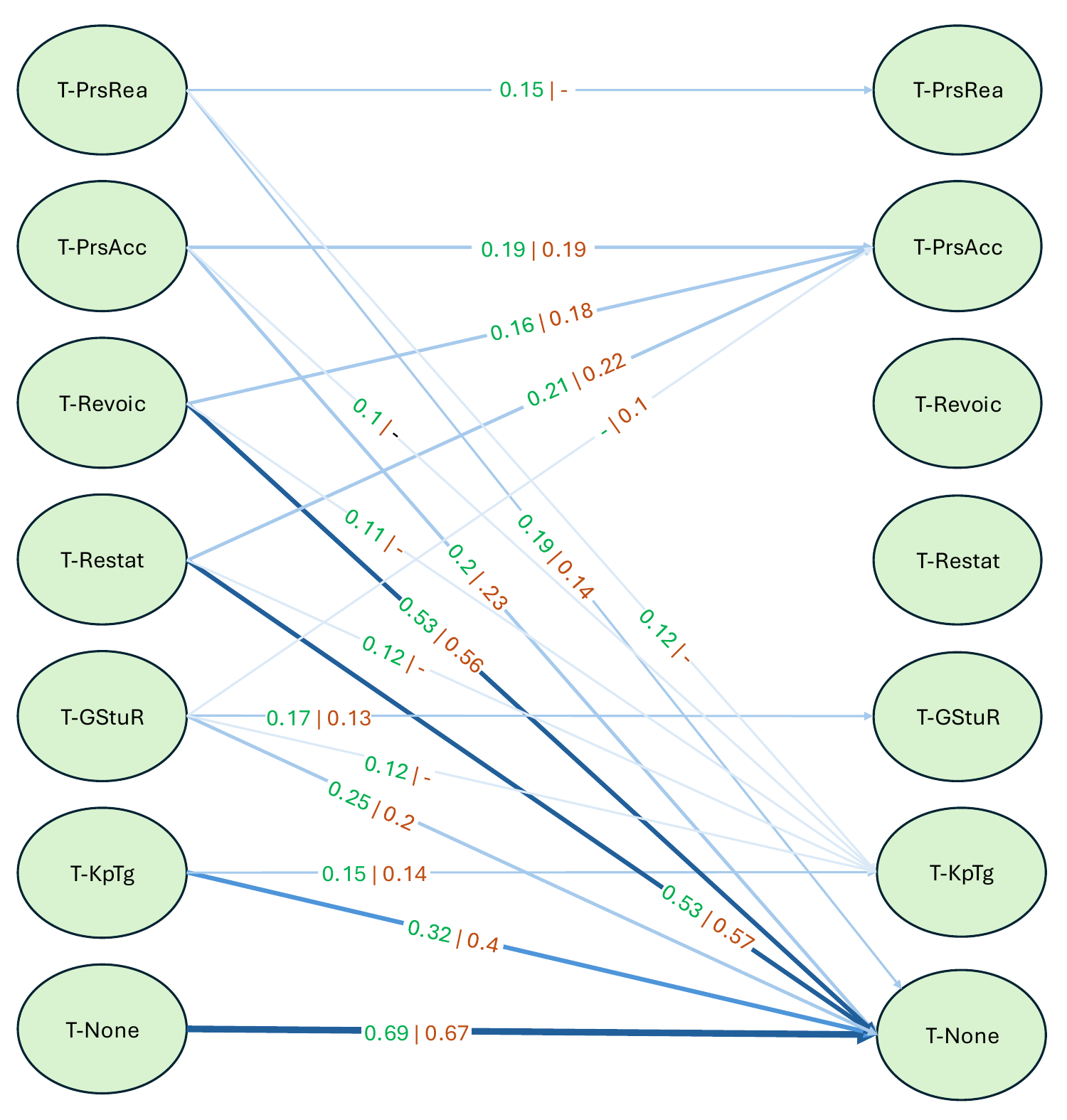}
            \caption{Intra Teacher Talk Move Transition}\label{fig:TT_talkmove_refined}
        \end{subfigure}
         \hfill
        \begin{subfigure}{0.47\textwidth}
            \centering
            \Description{This is an image of transition diagram depicting the transition probability from a student talk move to a teacher talk move.}
            \includegraphics[page = 3, width=\columnwidth]{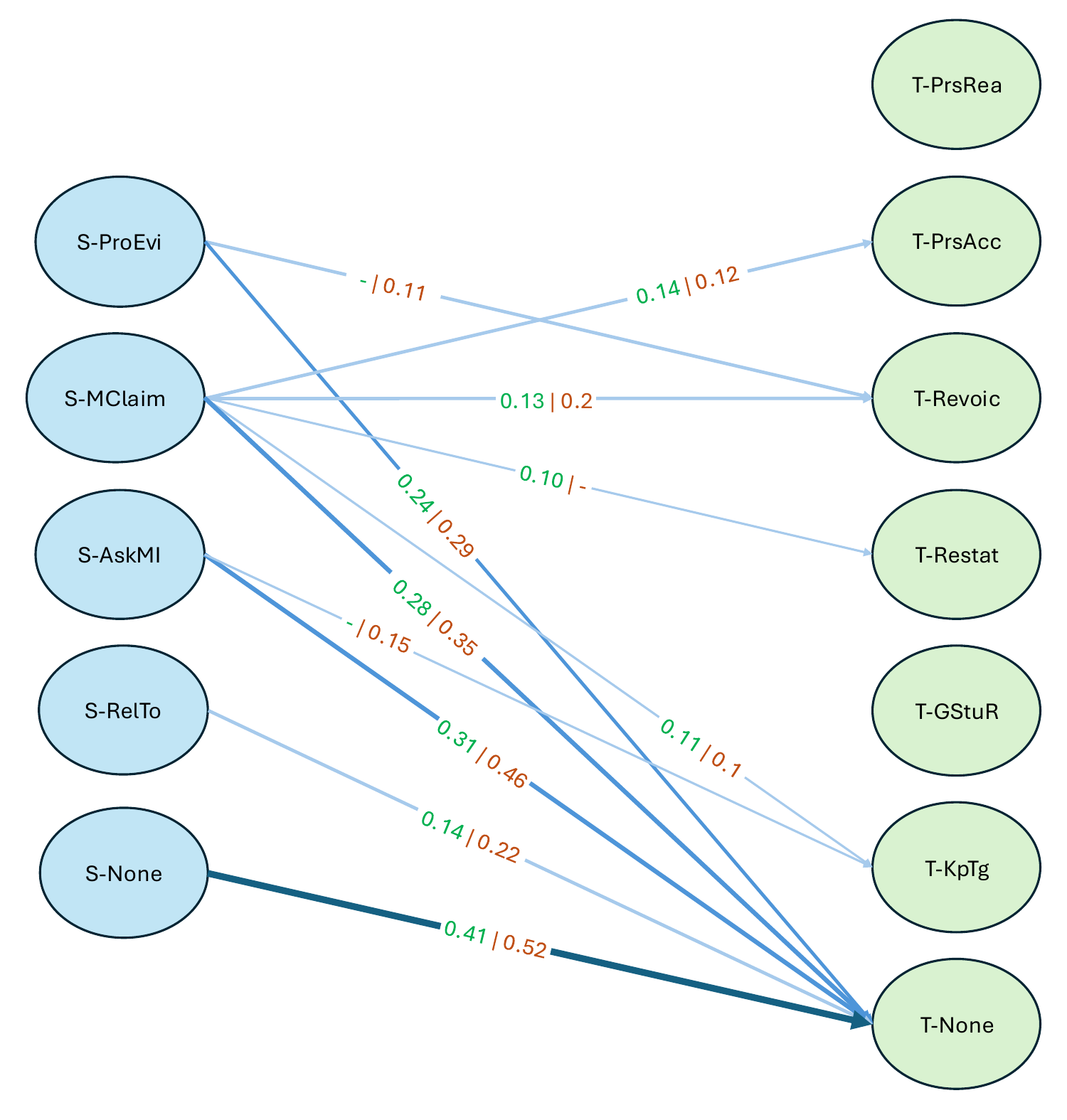}
            \caption{Student-Teacher Talk Move Transition}\label{fig:ST_talkmove_refined}
            \end{subfigure}
    \caption{Transition probabilities of one talk move being followed by another teacher talk moves in both the Teaching (numbers in green) and Tutoring (numbers in orange) dataset. The thickness of the edges represents the likelihood of transitions, with only probabilities above 10\% shown for clarity.}
    \label{fig:teahcer_trans_refined}
\end{figure}

\begin{figure}[!htb]
    \centering
    \begin{subfigure}{0.47\textwidth}
            \centering
            \Description{This is an image of transition diagram depicting the transition probability from a teacher talk move to a student talk move.}
            \includegraphics[page = 3,width=\columnwidth]{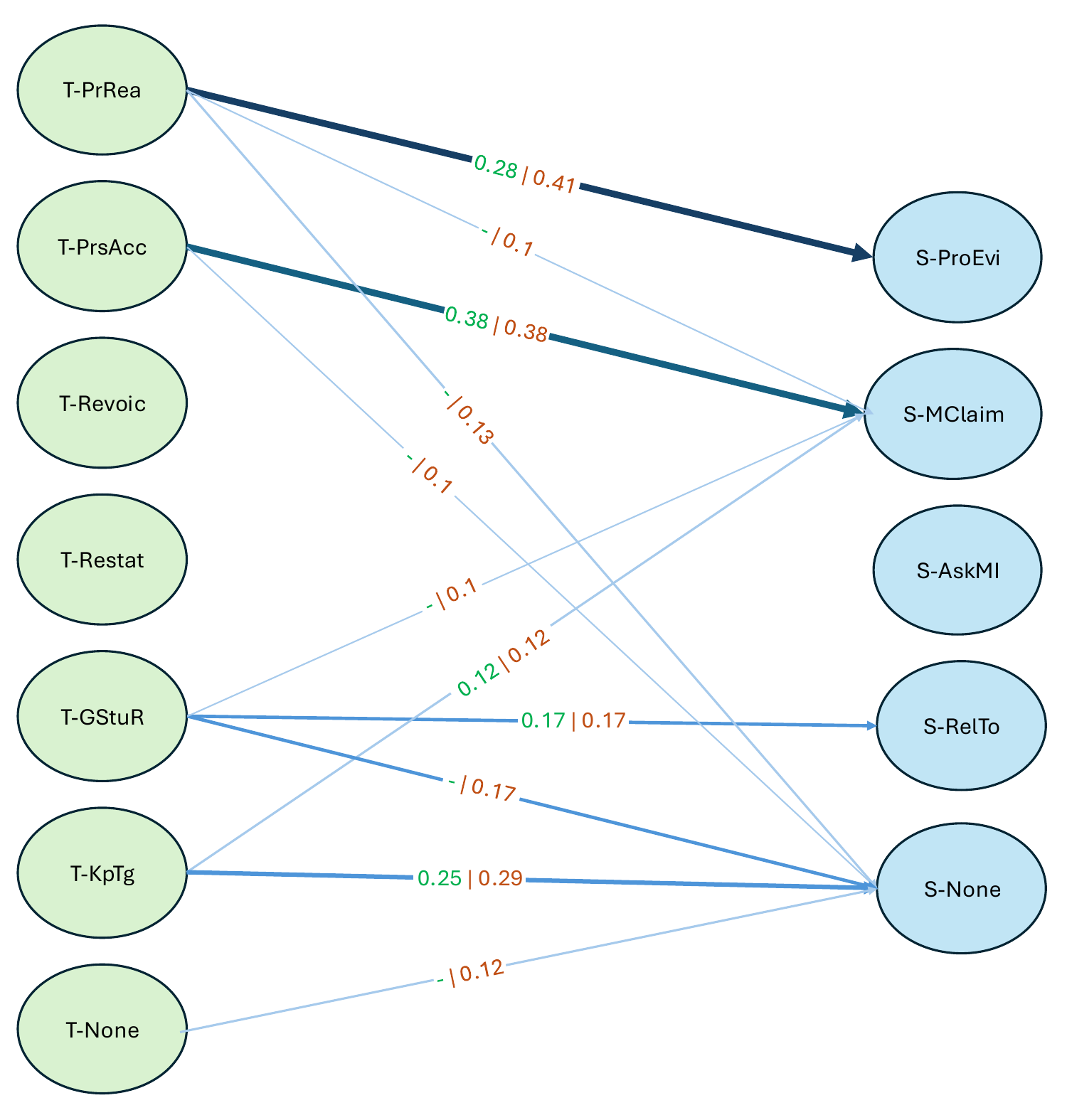}
            \caption{Teacher-Student Talk Move Transition}
            \label{fig:TS_talkmove_refined}
        \end{subfigure} 
        \hfill
        \begin{subfigure}{0.47\textwidth}
            \centering
            \Description{This is an image of transition diagram depicting the transition probability between two student talk moves.}
            \includegraphics[page = 3, width=\columnwidth]{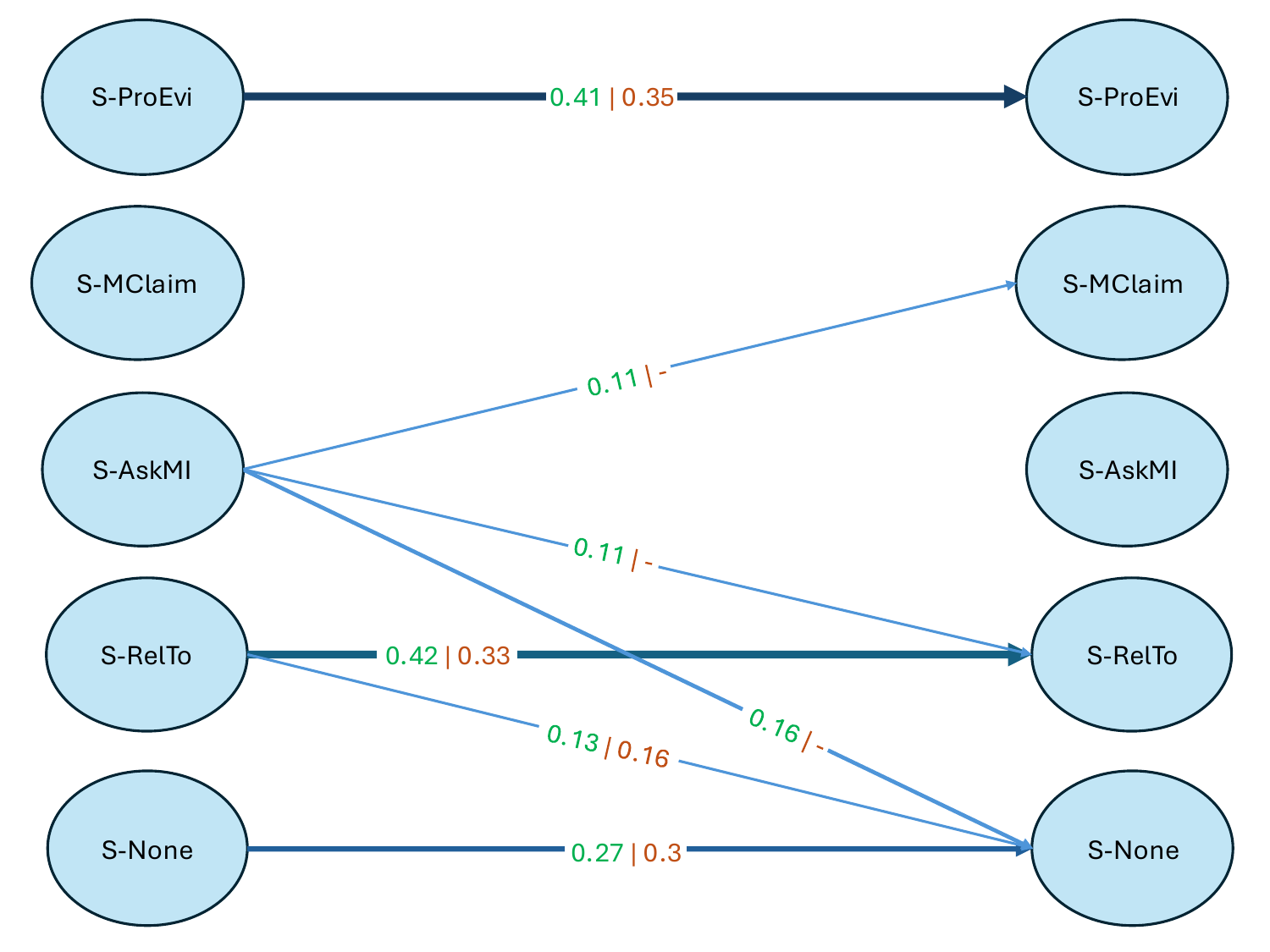}
             \caption{Intra Student Talk Move Transition}
            \label{fig:SS_talkmove_refined}
        \end{subfigure}
    \caption{Transition probabilities of one talk move being followed by another student talk move in both the Teaching (numbers in green) and Tutoring (numbers in orange) dataset. The thickness of the edges represents the likelihood of transitions, with only probabilities above 10\% shown for clarity.}
    \label{fig:student_trans_refined}
\end{figure}

Next we expanded our unigram analysis to bigram analysis to capture frequently occurring pedagogical behaviors in both the teacher and tutoring domains. To examine the interaction pattern between talk moves (and utterances without talk moves), we determined the transition probability between any two utterances. Figure~\ref{fig:teahcer_trans_refined} focuses on the teacher or tutor's response after a given talk move, while Figure~\ref{fig:student_trans_refined} focuses on the students' behavior after a talk move - either their own or made by other students or the teacher/tutor. The two transition probabilities on each edge in these figures represent the likelihood of one talk move being followed by another in teaching and tutoring respectively. Only transitions with a probability greater than 10\% are displayed. This 10\% threshold allows us to filter and focus on highly frequent instructional behaviors.

\paragraph{Transitions to Teacher/Tutor} Figure~\ref{fig:TT_talkmove_refined} shows that talk move pairs with the higher transition probabilities are consistent across both the teaching and tutoring datasets in the case of teacher-teacher and tutor-tutor talk move pairs. However, in the teaching dataset, all talk moves except for the \TNONE~ had a transition probability of 10\% or higher to the \TKET, whereas this pattern does not hold in the tutoring dataset. This discrepancy aligns with the notable (~3\%) difference in the occurrence of the \TKET talk move between the two datasets, indicating that well-trained teachers may provide more group engagement in the classroom. As shown in Figure~\ref{fig:ST_talkmove_refined}, all student talk moves are often followed by \TNONE, which further demonstrate that important information resides in non-talk move utterances, especially for those replying to \SAMI, \SMAC, and \SPRE. Additionally, the student talk move \SMAC are often followed by \TPRA, \TREV and \TKET in both domains, which highlights the potential need to suggest an instructional strategies based on the detailed content of students' claim. Furthermore, the observed transitions suggest a missed opportunity for educators to deepen student reflection and understanding by following a student’s \SMAC talk move with \TPRR.

\paragraph{Transitions to Students}
As shown in Figure~\ref{fig:TS_talkmove_refined}, the conceptually aligned talk move pairs to students: \TPRR → \SPRE, \TPRA → \SMAC, and \TGSR → \SRAS generally have high and similar transition probabilities across both the datasets. However, the transition probability for \TPRR → \SPRE pair is significantly higher in the tutoring dataset (41\%), exceeding that of the teaching dataset by 13\%. The pair \TKET → \SNONE also have moderately high transition probabilities in both datasets. Moreover, tutor talk moves show a slight tendency to be followed by student utterances without talk moves (\SNONE), consistent with the higher percentage of \SNONE-tagged student utterances in the tutoring dataset. As illustrated through Figure~\ref{fig:SS_talkmove_refined}, student-student talk move pairs also have almost similar transition probabilities across both the datasets, though the teaching domain exhibits a greater variety of transition pairs.


\subsubsection{Non-talk moves Utterances Interaction}
\label{sssec:rst-nonstrategic-transition}
\begin{figure}[t]
    \centering
    \begin{subfigure}{0.45\textwidth}
        \centering
         \Description{This is an image showing a heat map depicting the transition probability between talk moves excluding intermediary \textit{None} utterances in the teaching domain.}
        \includegraphics[width=\textwidth]{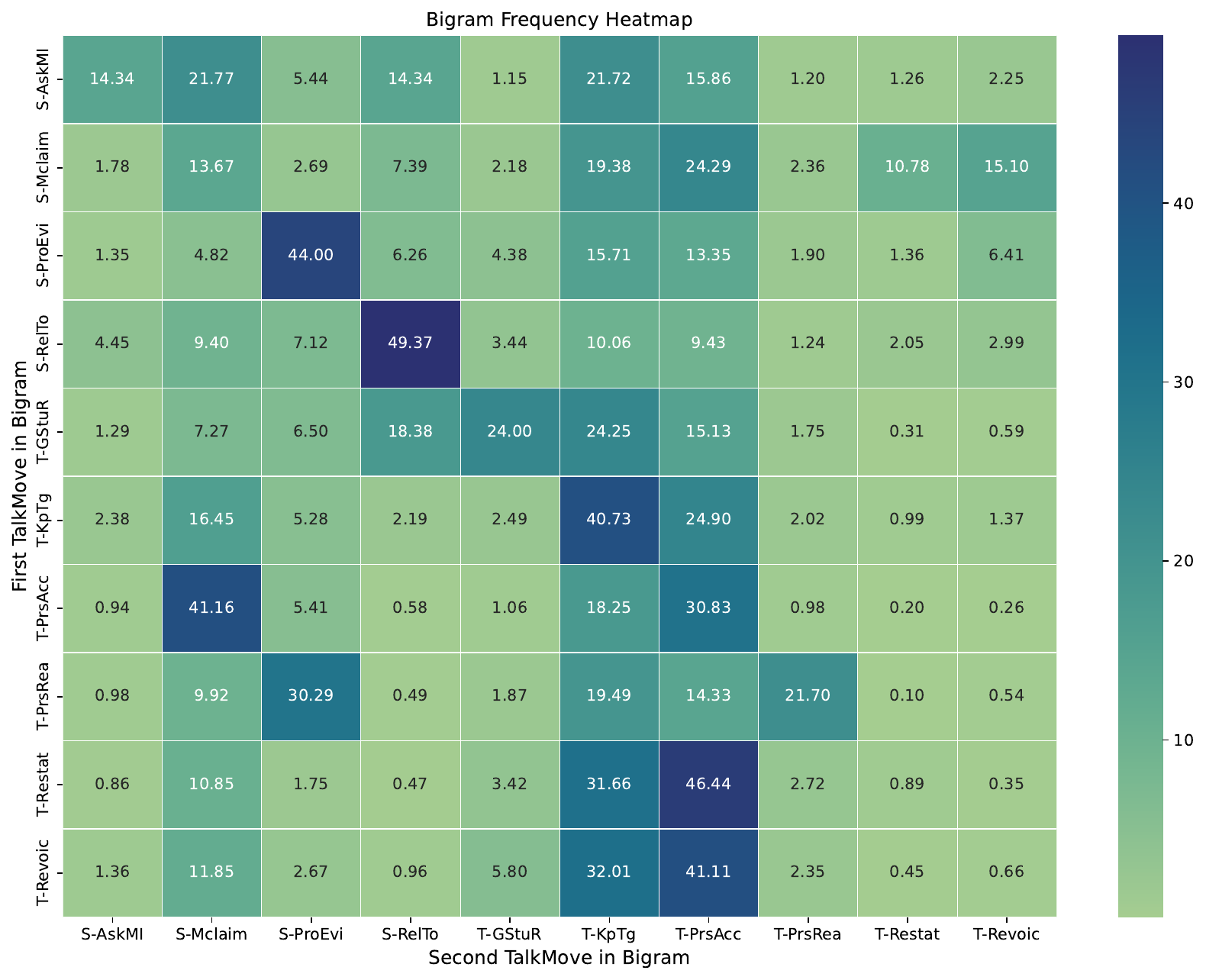}
        \subcaption{Teaching}
        \label{fig:teaching_app}
    \end{subfigure}
    \hfill
    \begin{subfigure}{0.45\textwidth}
        \centering
        \Description{This is an image showing a heat map depicting the transition probability between talk moves excluding intermediary \textit{None} utterances in the tutoring domain.}
        \includegraphics[width=\textwidth]{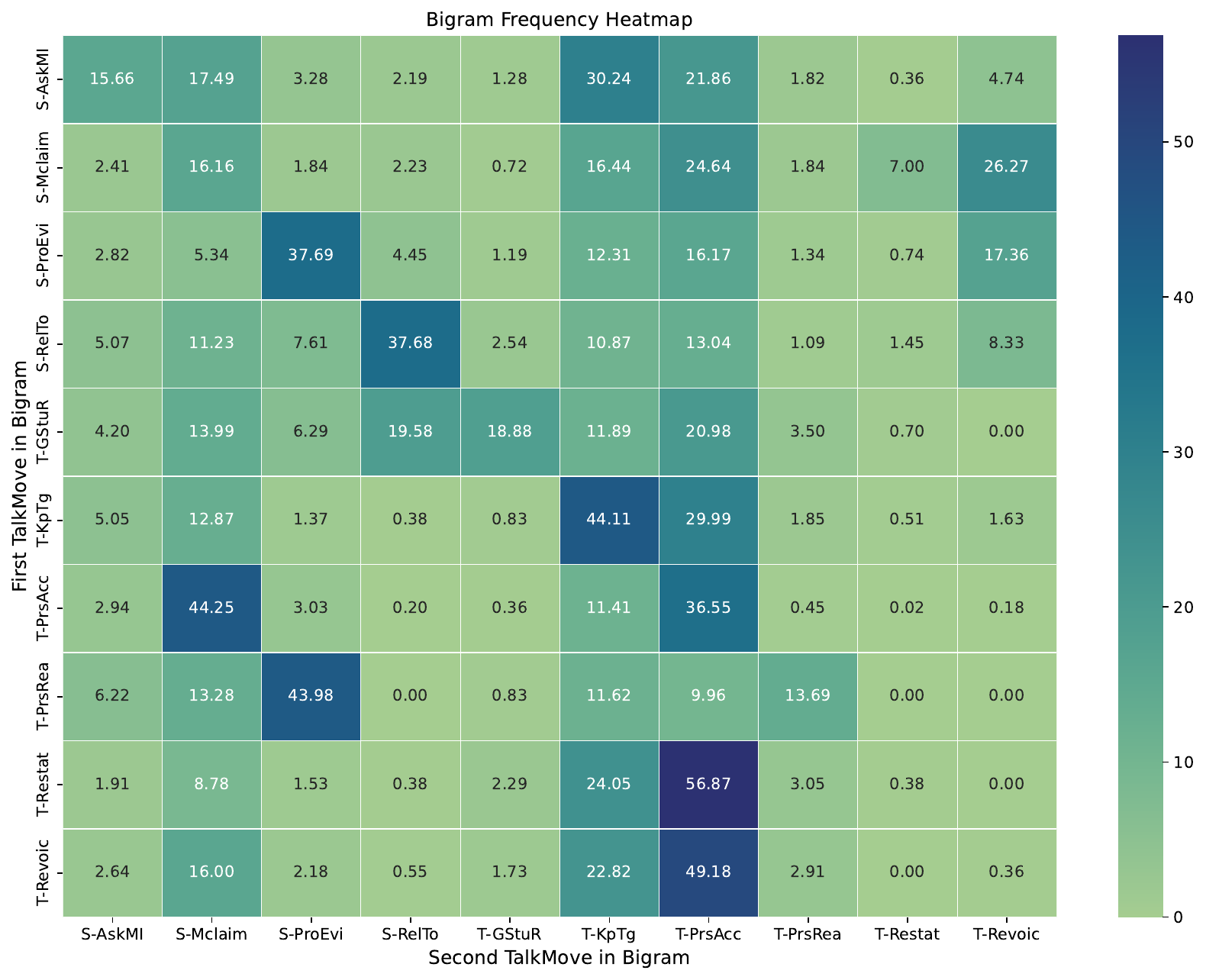}
        \subcaption{Tutoring}
        \label{fig:saga_app}
    \end{subfigure}
    \caption{Transition between Talk Moves excluding Intermediary \textit{None} Talk Moves}
    \label{fig:nonecollapsed_transition_app}
\end{figure}

\begin{figure}[t]
    \centering
    \begin{subfigure}{0.45\textwidth}
        \centering
        \Description{This is an image showing a heat map depicting the probability of \TNONE~ occurrence in between Talk Moves in the teaching domain.}
        \includegraphics[width=\textwidth]{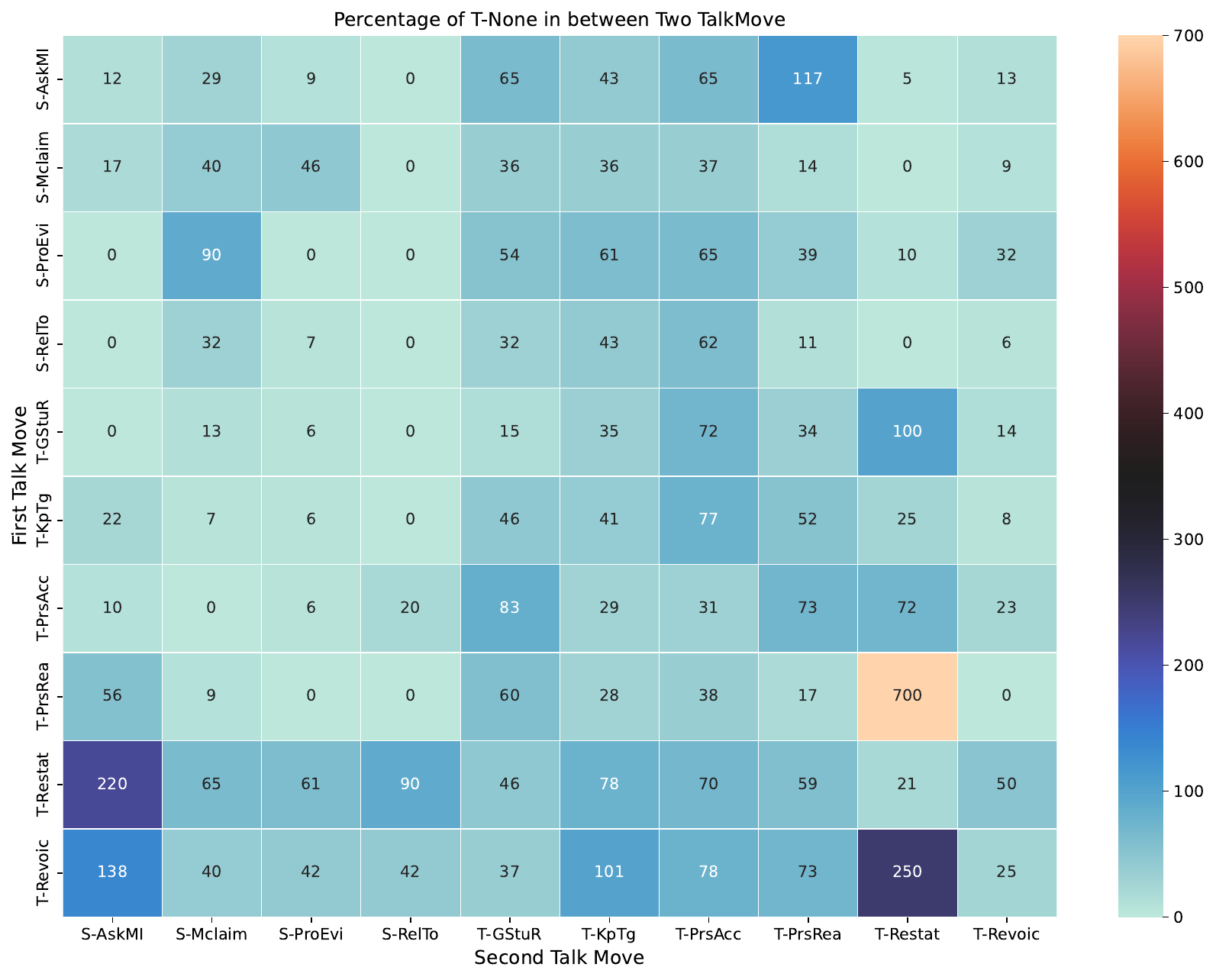}
        \subcaption{Teaching}
        \label{fig:nonecountTalkmove_app}
    \end{subfigure}
    \hfill
    \begin{subfigure}{0.45\textwidth}
        \centering
        \Description{This is an image showing a heat map depicting the probability of \TNONE~ occurrence in between Talk Moves in the tutoring domain.}
        \includegraphics[width=\textwidth]{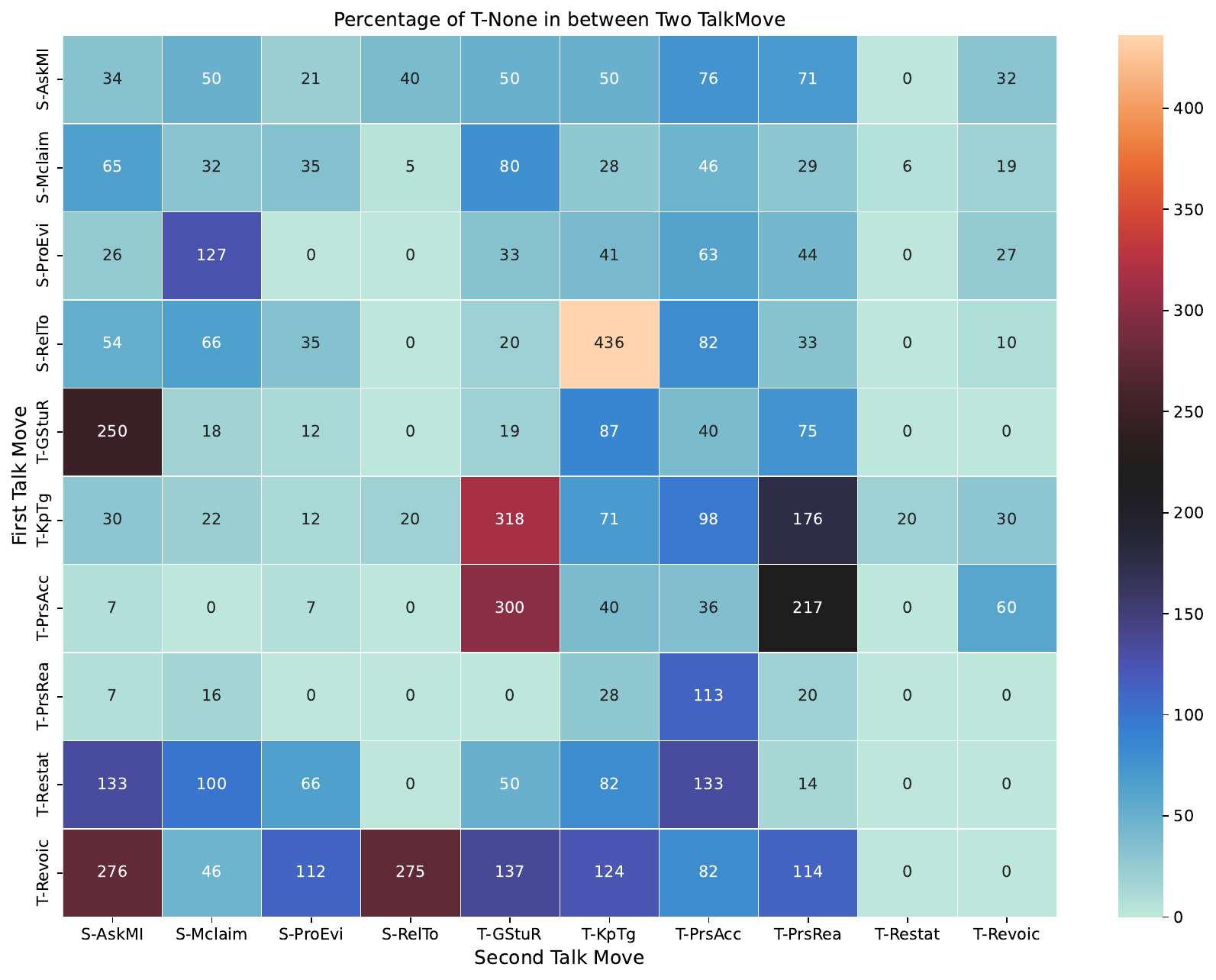}
        \subcaption{Tutoring}
        \label{fig:nonecountsaga_app}
    \end{subfigure}
    \caption{Comparison of Probability of \TNONE~ occurrence in between Talk Moves.}
    \label{fig:nonecountoverall}
\end{figure}

From the unigram distribution in Figure~\ref{fig:unigram} and the transition diagrams in Figure~\ref{fig:teahcer_trans_refined} and Figure~\ref{fig:student_trans_refined}, we can observe that interactions that don't involve talk moves (\TNONE) are the most prevalent in classroom discourse across teaching and tutoring domains. These interactions also have higher incoming transition rates from the talk moves. According to the procedure described in Section~\ref{ssec:method-seq-analysis}, Figure~\ref{fig:nonecountsaga_app} shows that the tutoring domain contains more \TNONE~interactions between talk move pairs compared to the teaching domain. In both datasets, self-transitions (indicated by the lighter shades along the diagonal in the heatmaps) have fewer \TNONE~interactions between them than other transitions~(see results on teaching domain in Figure~\ref{fig:nonecountTalkmove_app}). Additionally, \TRES and \TREV exhibit a higher occurrence of \TNONE~following them before transitioning to another talk move. Figure~\ref{fig:saga_app} shows a bigram analysis exclusively on talk moves by filtering out non-talk moves from the sequence of utterances in the tutoring settings. \TKET and \TPRA have relatively high incoming transition~(dark columns) indicates they are two frequent structural strategies without considering the None talk moves. Similar patterns are also shown in the teaching sessions~( Figure~\ref{fig:teaching_app})



In summary, the above transition diagram highlights high-frequency behavior patterns in talk move pairs. The probability distribution of \TNONE~separating talk-move pairs helps estimate the influence of other behaviors between these interactions. 


\subsection{Results from Multi-View Deep-Dive}
\label{ssec:rst-multi-view}
To gain deeper insights into the interactions uncovered in our previous sequential analysis~{\S\ref{ssec:rst-seq-talkmove}}, we turn to a multi-view analysis via discourse relations considering the transitions among utterances with and without talk moves.

\begin{table}[!htp]
    \centering
    \setlength{\tabcolsep}{1pt}
    \caption{Comparison of Bigrams in Teaching and Tutoring Datasets}
    \begin{tabular}{lccclcccl}
        \toprule
        \multirow{2}{*}{Bigrams} & \multicolumn{2}{c}{Teaching Dataset} & \multicolumn{2}{c}{Tutoring Dataset} \\
        \cmidrule(lr){2-3} \cmidrule(lr){4-5}
        & \thead{Trans \\ Prob} & \thead{Discourse \\ Relations} & \thead{Trans \\ Prob} & \thead{Discourse \\ Relations} \\
        \midrule

         \thead{\SMAC ~-\\ \TPRA} & 14\% & \thead{ClariQ.(21.1\%)\\ Cont. (19.69\%)\\ QElab.(10.49\%)}  & 12\% & \thead{ClariQ.(33\%)\\ Cont. (8.37\%)\\ QElab.(6.65\%)} \\

         \thead{\SMAC ~-\\ \TREV} &  13\% & \thead{Cont.(33.53\%)\\ Elab. (14.56\%)\\ Ack.(11.22\%)} &   20\% &  \thead{Ack.(31.60\%)\\ Corr. (14.88\%)\\ Cont.(10.28\%)}   \\
         
        \thead{\SRAS ~-\\ \SRAS} & 42\% & \thead{Cont. (19\%)\\ Corr. (10.18\%)\\ Ack. (8.32\%)}  & 33\%  & \thead{Cont. (11.96\%)\\ Corr. (7.61\%)\\ Ack. (7.61\%)} \\
        
        \thead{\SPRE ~-\\ \SPRE} & 41\% & \thead{Elab. (46.81\%)\\ Cont. (40.51\%)\\ Corr. (3.10\%)\\ Contr. (3.01\%)} & 35\% & \thead{Cont. (28.57\%)\\ Elab. (21.43\%)\\ Corr. (6.30\%)}\\
        
        
        \bottomrule
    \end{tabular}
    \label{tab:bigramDR}
\end{table}
\subsubsection{Discourse Relations involving Talk Moves}
\label{ssec:strategic-discourse}

\begin{figure}[tb]
    \centering
    \Description{This is an image showing examples of a couple of talk move transition with the discourse relation in between them.}
    \includegraphics[width=1\columnwidth]{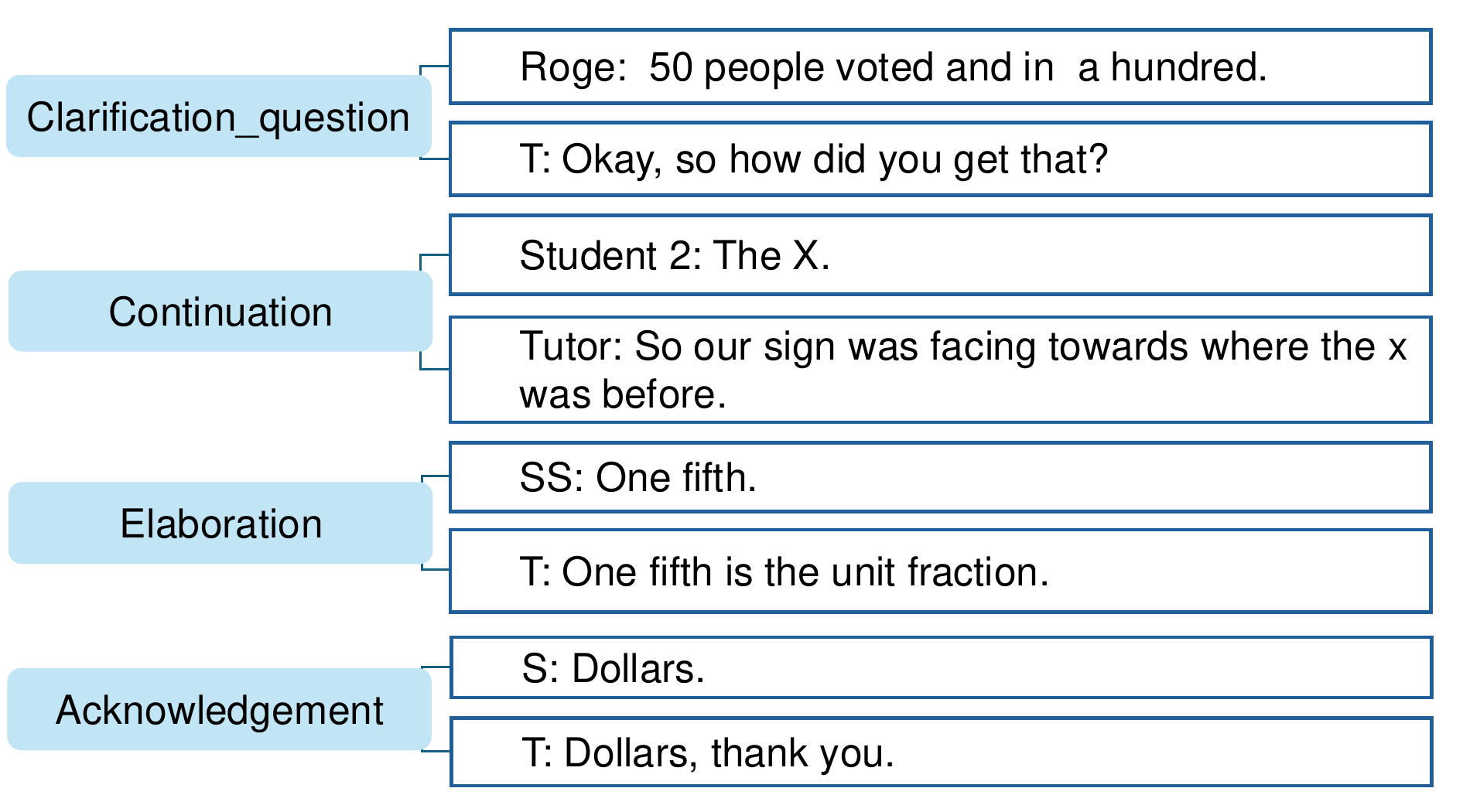} 
    \caption{Example of the \SMAC → \TPRA talk move pair with the \textit{Clarification Question} discourse relation and the \SMAC → \TREV talk move pair with the \textit{Continuation, Elaboration}, and \textit{Acknowledgement} discourse relations.}
    \label{fig:mcrevprs}
\end{figure}

Table~\ref{tab:bigramDR} highlights the key findings from our analysis of discourse relations among talk moves. The first column `Bigram' shows the talk move pairs detected in our previous sequential talk move analysis. We selected 4 talk move pairs with a relatively high transitional probability (`Trans Prob') from the teaching and tutoring datasets. The `Discourse Relations' column shows the portion of the top 3 discourse relations between each pair of the talk moves. We observe from this table that when replying to a students' claim, an actionable suggestion for a teacher or tutor could be \TPRA via asking a clarification question or simply revoicing the claim with some continuation or elaboration~(Figure~\ref{fig:mcrevprs}). 

\begin{figure}[!htb]
    \centering
    \Description{This is an image showing examples of the \SRAS - \SRAS talk move pair with the \textit{Continuation}, \textit{Contrast}, and \textit{Acknowledgement} discourse relations.}
    \includegraphics[width=1\columnwidth]{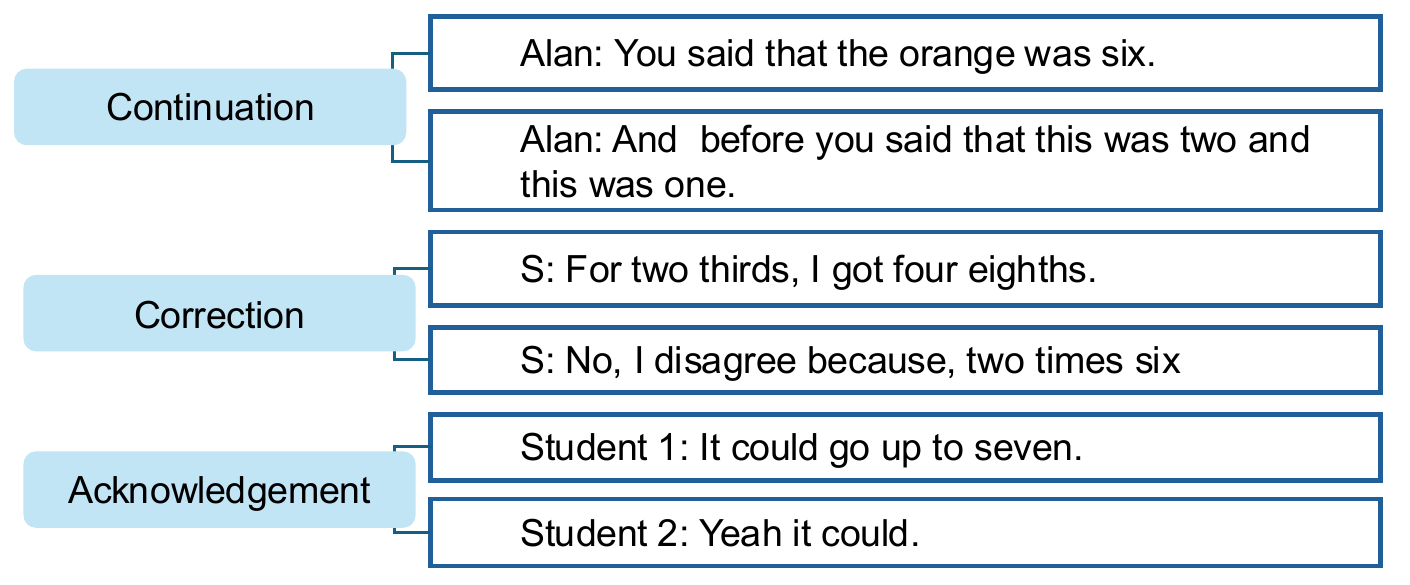} 
    \caption{Examples of the \SRAS - \SRAS with the \textit{Continuation}, \textit{Contrast}, and \textit{Acknowledgement} discourse relations.}
    \label{fig:rasras}
\end{figure}

\begin{figure}[!htb]
    \centering
    \Description{This is an image showing examples of the \SPRE - \SPRE talk move pair with the \textit{Elaboration}, \textit{Continuation}, \textit{Correction}, and \textit{Contrast} discourse relations.}
    \includegraphics[width=1\columnwidth]{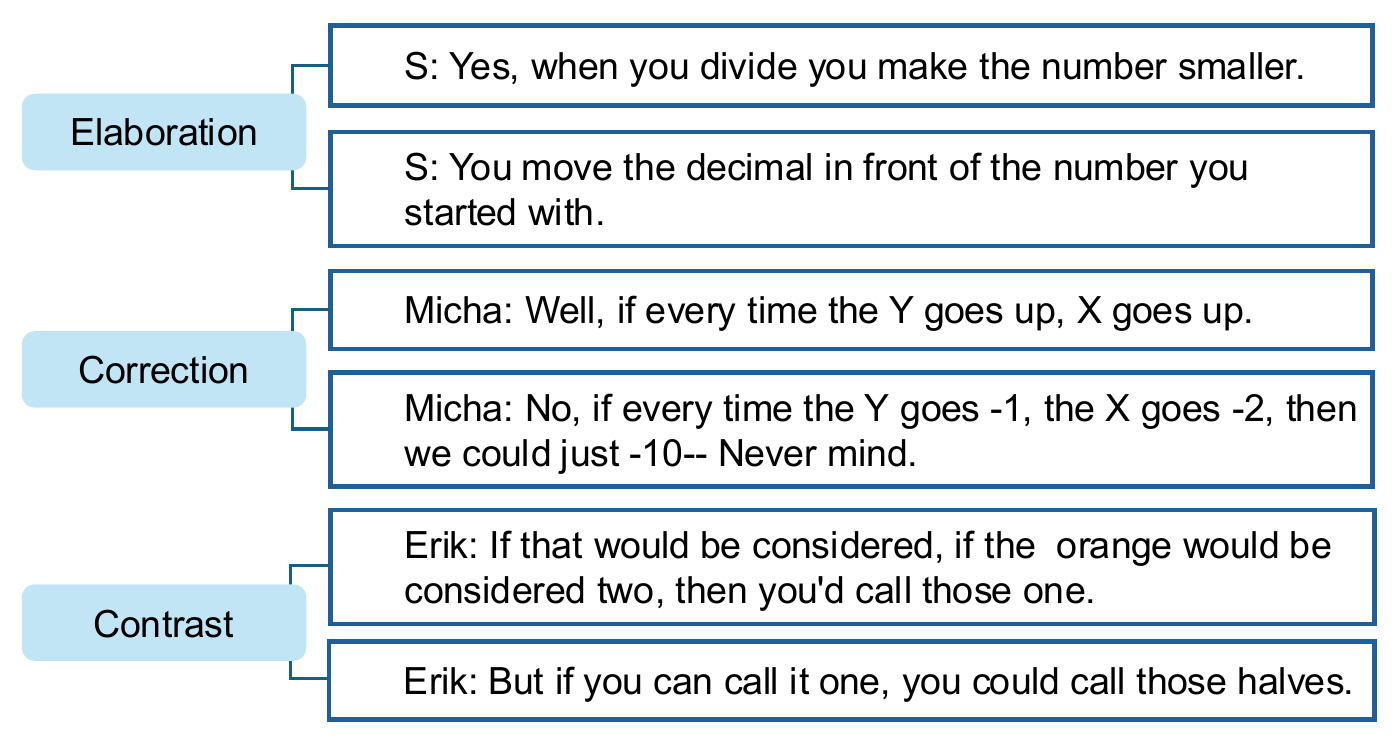} 
    \caption{Examples of the \SPRE - \SPRE talk move pair with the \textit{Elaboration}, \textit{Continuation}, \textit{Correction}, and \textit{Contrast} discourse relations.}
    \label{fig:prsevi}
\end{figure}

When designing a student bot to collaborate with students as they learn mathematics, developers seek to simulate the behaviors of a student, for example as they relate other students or provide evidence across multiple utterances. Figure~\ref{fig:rasras} and Figure~\ref{fig:prsevi} elucidates instances of these types of talk move pairs. Further lexical analysis shows that 22\% of the talk move pairs in the teaching dataset and 32\% in the tutoring dataset use the conjunction word "so" to connect continuous utterances.

\subsubsection{Importance of Utterances without Talk Moves}
\label{sssec:non-stragic-disource-relation}
We conducted an in-depth analysis of discursive interactions not involving talk moves using two approaches. First, we examined bigram pairs consisting of one talk move and one non-talk move utterance. Second, we analyzed instances where utterances classified as None occured between two talk moves that exhibit significant transition probabilities given our prior analyses. The frequent DAs associated with these significant transition patterns are detailed in Table~\ref{tab:talkmoveDA}.

\begin{figure}[!htb]
    \centering
     \Description{This is an image showing examples of \TRES → \TNONE~ transition with different discourse relations and DAs.}
    \includegraphics[width=1\columnwidth]{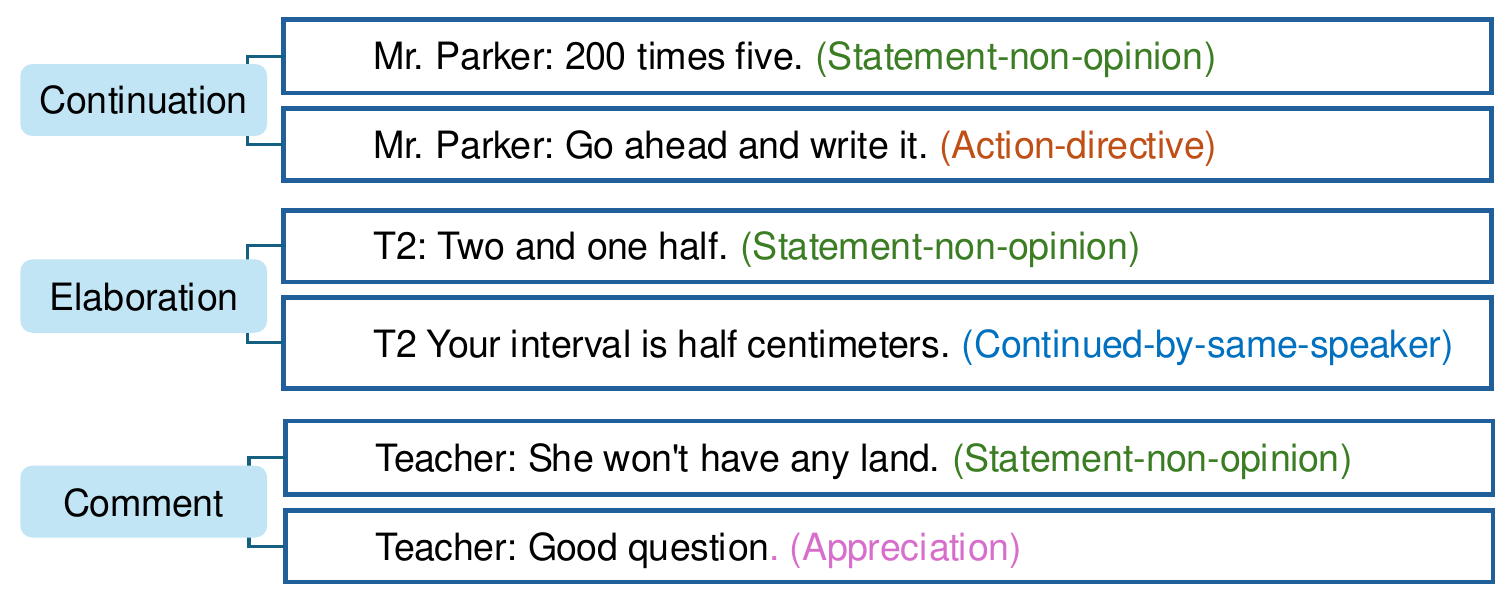} 
    \caption{Examples of the \TRES → \TNONE~ pair with different discourse relations and DAs.}
    \label{fig:restatnone}
\end{figure}

\paragraph{Talk Moves $\rightarrow$ None}
The teacher talk moves \TRES, \TKET, and \TPRA exhibit high transition rates towards utterances classified as \TNONE in both the teaching and tutoring domains, as observed in Figure~\ref{fig:TT_talkmove_refined}. The predominant discourse relations for these transitions are \textit{Continuation}, \textit{Elaboration}, and \textit{Comment}. The most frequent DAs associated with \TNONE s in this pair are presented in Table~\ref{tab:talkmoveDA}. Figure~\ref{fig:restatnone} presents examples of the \TRES → \TNONE~ pair, showcasing various discourse relations and DAs. From this figure, we can observe that utterances classified as \TNONE~ can actively contribute to the classroom discourse by introducing new information, guiding the flow of conversation, or acknowledging and appreciating student contributions. These acts can play a crucial role in fostering productive classroom dialogue. 

Similarly, the frequently associated DAs and discourse relations with the pair \TKET → \TNONE~suggests that the purpose of the \TKET in these examples is to draw the attention of students and encourage active listening.  Utterances classified as \TNONE following \TKET talk moves may provide students with more direction, helping to better engage and understand the flow of the discussion. Our analysis also reveals that the \TNONE~talk move in the \TPRA → \TNONE~can support \TPRA by offering additional information and directions, ensuring that a question is clearly conveyed and effectively understood by the students. 

\begin{figure}[!htb]
    \centering
    \Description{This is an image showing examples of the same-category talk moves separated by a teacher utterance classified as \TNONE.}
    \includegraphics[width=1\columnwidth]{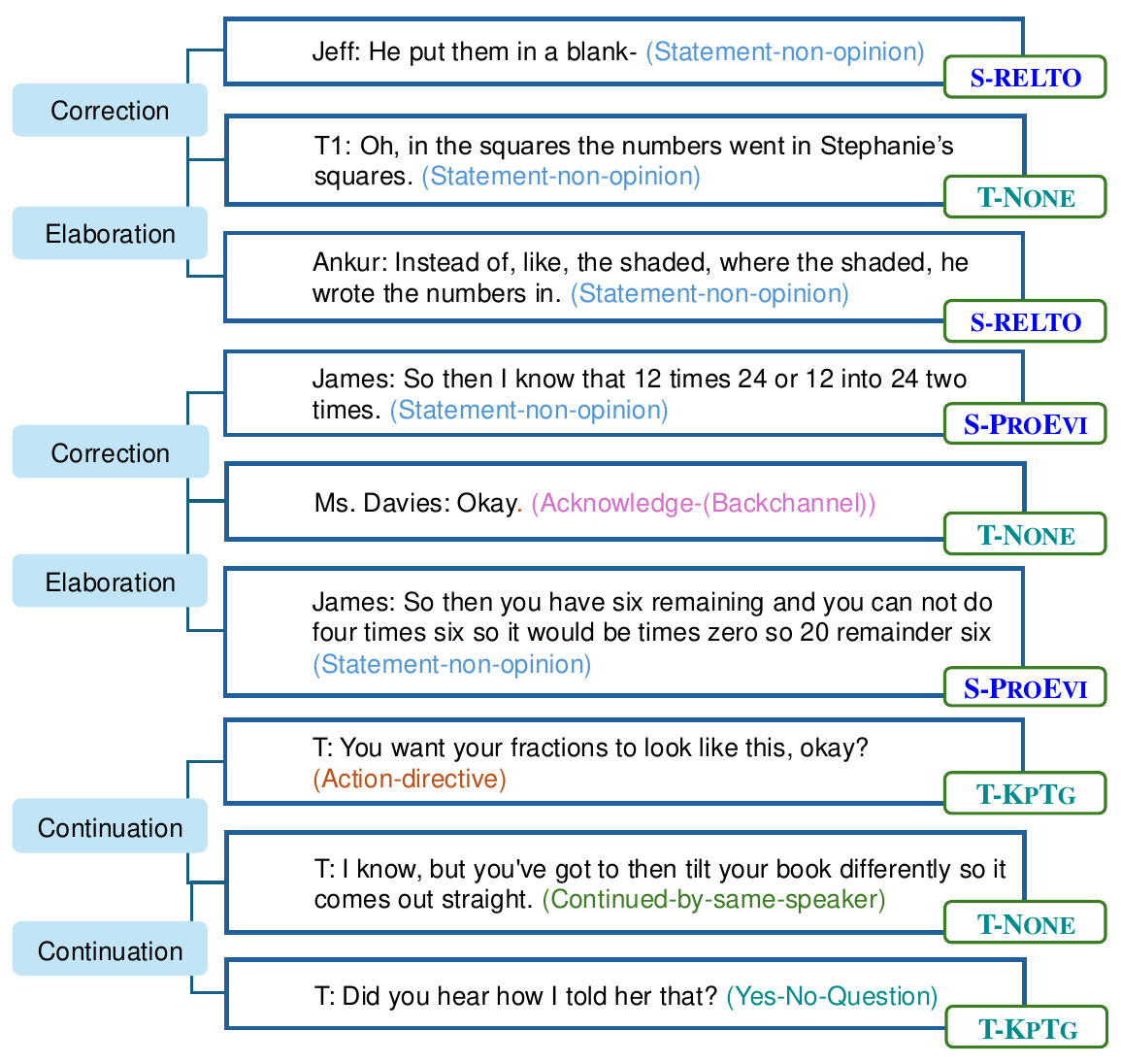} 
    \caption{Examples of the same-category talk moves separated by a teacher utterance classified as \TNONE.}
    \label{fig:samepairtnone}
\end{figure}

\paragraph{Talk Moves $\rightarrow$ None $\rightarrow$ Talk Moves}
We further analyzed the role of (\TNONE) utterances occurring between two consecutive talk moves, focusing first on cases where the talk moves belong to the same category. Based on the qualitative examples illustrated in Figure~\ref{fig:samepairtnone} and the importance of these transitions supported by Table~\ref{tab:talkmoveDA}, we can see that \TNONE~between same-category student talk moves help to acknowledge, correct, and guide students, enabling them to refine their understanding while encouraging further contributions to the classroom discourse. Meanwhile, \TNONE~between same-category teacher talk moves can act as a bridge, seamlessly linking the two moves and creating the effect of a multi-utterance spanning talk move.

\begin{figure}[!htb]
    \centering
    \Description{This is an image showing examples of different-category talk moves separated  by a teacher utterance classified as \TNONE.}
    \includegraphics[width=1\columnwidth]{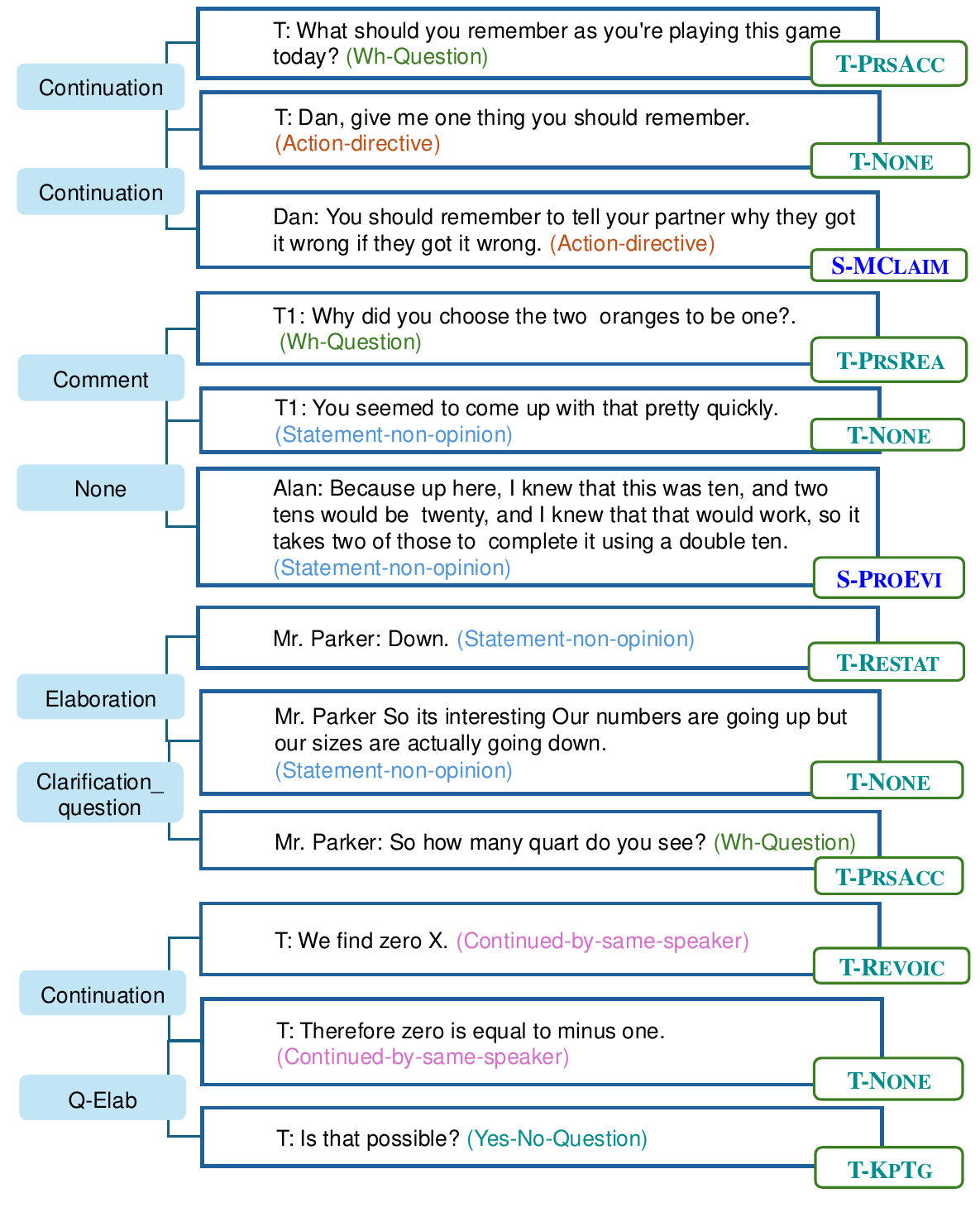} 
    \caption{Examples of different-category talk moves separated  by a teacher utterance classified as \TNONE.}
    \label{fig:diffpairtnone}
\end{figure}

When exploring the role of teacher utterances classified as \TNONE 
in transitions between different talk moves, we observe that these utterances help with elaborating on student contributions, guiding discourse, and fostering engagement. Additionally, \TNONE~ utterances can strengthen the coherence of teacher talk moves, ensuring a fluid and connected sequence of strategic interactions. Illustrative examples of these transitions are presented in Figure~\ref{fig:diffpairtnone}.

\begin{table}[t]
    \scriptsize
    \centering
    \renewcommand{\arraystretch}{1.3} 
    \caption{Dialogue Acts Associated with Non-Talk Moves in Interaction with Talk Moves. (Dialogue Acts: ContS. = Continued-by-same-speaker, StatNO = Statement-non-opinion, AcD. = Action-directive, AckB. = Acknowledge (Backchannel), R-Ack. = Response-Acknowledgement, Apr. = Appreciation, StatO. = Statement-opinion, Thanking = Thanking, Other = Other).}
    \begin{tabularx}{0.47\textwidth}{p{.3\linewidth} p{.55\linewidth}} 
    \hline
    \textbf{TalkMove Pairs} & \textbf{Dialogue Acts (of the intervening \TNONE)} \\
    \hline
    \multicolumn{2}{c}{\textbf{Talk Move to Non-Talk Move Transition}} \\
    \hline
    \TRES → \TNONE & ContS. (43.74\%), StatNO (19.88\%), AcD. (8.39\%), AckB. (7.29\%), Apr. (3.76\%) \\

    \TKET → \TNONE & StatNO (26.65\%), ContS. (24.05\%), AcD. (19.43\%), StatO. (4.94\%) \\
   
   \TPRA → \TNONE & StatNO (35.93\%), ContS. (23.67\%), AcD. (20.34\%) \\
   
    \SMAC → \TNONE & ContS. (61.66\%), AckB. (14.83\%), R-Ack. (5.69\%), StatNO (4.18\%), Apr. (3.90\%) \\
  
    \SPRE → \TNONE & ContS. (54.62\%), AckB. (23.15\%), R-Ack. (4.57\%), Apr. (4.52\%), StatNO (4.11\%) \\
    
    \SRAS → \TNONE & AckB. (14.50\%), R-Ack. (7.14\%), StatNO (5.09\%), AcD. (4.00\%) \\
   \hline
    \multicolumn{2}{c}{\textbf{Intra-Talk Move Transition with Intervening Non-Talk Move}} \\
   \hline
    \SRAS → \TNONE → \SRAS & ContS. (55.17\%), AckB. (17.24\%), StatNO (17.24\%), R-Ack. (3.45\%) \\
    
    \SPRE → \TNONE → \SPRE & ContS. (47.47\%), AckB. (33.33\%), StatNO (6.06\%), AcD. (5.05\%) \\
    
    \TKET → \TNONE → \TKET & StatNO (27.28\%), ContS. (24.30\%), AcD. (17.48\%), Other (6.24\%), StatO. (5.28\%) \\
   
    \TPRA → \TNONE → \SMAC & StatNO (39.35\%), ContS. (29.60\%), AcD. (14.80\%), StatO. (3.25\%) \\
    
    \TPRR → \TNONE → \SPRE & StatNO (28.57\%), AcD. (21.43\%), ContS. (21.43\%), Thanking (7.14\%) \\
  
    \TRES → \TNONE → \TPRA & ContS. (42.59\%), StatNO (16.67\%), AckB. (8.33\%), R-Ack. (8.33\%), AcD. (6.94\%) \\
    
    \TREV → \TNONE → \TKET & ContS. (39.81\%), StatNO (19.91\%), AcD. (13.89\%), AckB. (6.94\%), StatO. (3.70\%) \\
    \hline
    \end{tabularx}
    \label{tab:talkmoveDA}
\end{table}

\section{Discussion}
We analyzed pedagogical behaviors in two mathematics education datasets, \TB~(teaching) and \SAGA~(tutoring). Using manually annotated talk moves and two state-of-the-art models for dialogue act~(DAs) and discourse relation~(DRs) prediction, we conducted a top-down analysis on discursive behaviors looking at (1) unigram patterns, (2) sequential patterns, and (3) a deep dive via multi-view analysis.
\subsection{Main Findings}
\label{ssec:main-findings}
Our unigram analysis of utterance-level talk moves and dialogue acts revealed similar overarching distributional patterns across the teaching and tutoring datasets. However, the tutoring dataset~(\SAGA) exhibited a slightly higher prevalence of utterances without talk moves, with \TNONE~occurring 5.8\% more frequently and \SNONE 4.3\% more frequently relative to the teaching (\TB) dataset. This discrepancy may stem from the fact that well-trained teachers are more likely to employ talk moves, leading to a more structured and intentional discourse in classroom settings. A deeper joint analysis of DAs with talk moves provides a more nuanced perspective on discursive interactions and also helps us further differentiate between the teaching and tutoring datasets. While similar patterns were observed for utterances with and without talk moves in both domains,~\DACC played a more significant role in utterances without talk moves during the tutoring sessions. This highlights how the differences~(e.g., amounts of students, different training of the teachers/tutoring, and in-person vs online) between teaching and tutoring influences discourse flow, potentially altering communication dynamics. Moreover, the higher ratio of student talk moves followed by another student talk move in the teaching domain and the higher frequency of student talk moves followed by non-talk move utterances by the instructor in the tutoring domain may indicate that the teachers were more adept at fostering student-driven discussions than the tutors.  Additionally, the lower transition ratio to \TRES in tutoring sessions suggests a potential area for actionable feedback to the tutors.

Finally, through a detailed analysis of frequent behaviors, we identified meaningful discourse patterns in talk move sequences. We discovered that, beyond talk moves and dialogue acts, the dependency relations within the discourse provide valuable insights that can inform instructor feedback and guide action policies for future AI agents in mathematics education. Our analysis shows how students' discursive participation patterns may reveal important information about their needs and state of mind, enabling teachers and AI agents to adjust their strategies for more targeted support. Additionally, we uncovered patterns in teachers' behavior beyond their talk moves, such as their role as the primary action director in the classroom discourse and their use of various strategies "under the hood" of talk moves to address different situations.
Moreover, despite being not being categorized as include a talk move, \TNONE~can significantly contribute to classroom discourse by guiding discussions, acknowledging student input, and maintaining coherence in teacher talk moves. These utterances help to structure transitions, offering clarification and direction. Between student moves, they can play the role of acknowledging and refining ideas. Between teacher moves, they can act as a bridge and foster continuity. These findings underscore the nuanced role of talk moves and dialogue acts, offering insights to enhance teacher training and the development of intelligent tutoring systems.

\subsection{Limitations and Future Work}
\label{ssec:limitations}
A limitation of this work is that, due to a lack of annotated dialogue acts and discourse relation data, we only selected two of the available state-of-art schema with public accessible models, SWBD-DAMSL and SDRT for dialogue acts and discourse relation respectively. Designing the most suitable schema for educational dialogue still requires future investigation. Further more, the two off-the-shelf models on SWBD-DAMSL and SDRT were not fine-tuned in our mathematics education datasets \TB~and~\SAGA, which may lead to suboptimal results. Future work on annotating a larger amount of in-domain training data will enable potential finetuning or multi-task learning for jointly modeling all three tasks, which could further increase the accuracy of our proposed toolkits for the multi-perspective analysis. In this paper, our analysis mainly focused on educational theory-grounded talk moves analysis with both well-annotated datasets and finetuned models. We mainly focused on  bigram talk moves sequences with or without non-talk moves utterances in between. Jointly considering a connected multi-view sub-graph could lead to the discovery of other interesting patterns of pedagogical behaviors. Finally, we did not associate the dialog interactions with an analysis of the instruction quality or student learning outcomes. Future work combining these kinds of data may provide more insights into the nature of  high-quality instruction. 

\subsection{Potential Applications}
\label{ssec:potential-application}
This work offers an in-depth analysis of mathematics instructional dialogue in both the teaching and tutoring domains, which may enable educators and future AI agents to facilitate more effective discussions with students. Our empirical analysis suggests that dialogue acts and discourse relations could offer a more nuanced understanding of the behavioral patterns of students and teachers embedded within theory-driven talk moves. A multi-view analysis may lead to actionable feedback provided to educators that goes beyond information based solely on talk moves~\cite{jacobs2022promoting,jensen2020toward,Jensen.2020}, including important dialogue captured by~\TNONE~and~\SNONE utterances. Moreover, a deeper understanding of these dialogue structures can inform the development of AI-driven tutoring systems in mathematics education~\cite{bin2022artificial,latham_conversational_2022,tack2022ai}. By integrating these findings with advancements in generative AI and domain-specific datasets, this research paves the way for broader applications beyond mathematics. Potential areas of impact include collaborative learning~\cite{breideband2023community,cao2023comparative,cao2023designing,kim2022learning}, creative problem solving~\cite{boussioux2024crowdless,wu2021ai}, and fostering meaningful student engagement~\cite{huang2023effects}. This work lays the foundation for AI-driven educational tools that enhance explainable instructional quality, promote controllable learning experiences, and support the evolving needs of both educators and students.

\section{Conclusion}
 Integrating talk moves, dialogue acts, and discourse relations, our multi-perspective study reveals key insights into the nature of teaching and tutoring discourse across two mathematic education datasets: \TB and \SAGA.
Our analysis reveals that while teaching and tutoring datasets share overarching distributional similarities in their teacher and student talk moves, tutoring interactions exhibit a slightly higher prevalence of utterances not classified as containing a talk move, suggesting a need for targeted tutor training. Joint dialogue act analysis underscores the nuanced role of diverse dialogue acts in enhancing strategic communication.  Notably, transition analysis highlights tutors' greater reliance on utterances that do not contain a talk move and a reduced tendency to restate students' ideas, suggesting that actionable feedback in these areas might be appropriate. Furthermore, frequent teacher-student interaction patterns align with core educational clusters, emphasizing the structured nature of pedagogical discourse. Beyond utterance-level talk moves, discourse dependency relations offer insights for optimizing AI-driven educational agents. Our deeper exploration into utterances without talk moves reveals their essential function in shaping classroom dialogue. Rather than mere fillers, they can serve as pivotal elements in guiding, acknowledging, and structuring discourse. Whether linking teacher talk moves for coherence or scaffolding student contributions, utterances with and without talk moves likely play a crucial role in engagement and comprehension. These findings underscore the importance of integrating discourse relations and dialogue acts into AI-assisted education to foster more effective and responsive learning environments.

\section{Acknowledgments}
The authors would like to thank the anonymous reviewers for their valuable feedback. This research was supported by the National Science Foundation grant \#2222647 and the NSF National AI Institute for
Student-AI Teaming (iSAT) under grant DRL \#2019805. All opinions are those of the authors and do not reflect those of the funding agencies.

\bibliographystyle{abbrv}
\bibliography{bib-files/cl,bib-files/dialog-act,bib-files/discourse,bib-files/edm,bib-files/nlp}

\end{document}